\documentclass[lettersize]{IEEEtran}
\usepackage{amsmath,amsfonts}
\usepackage{algorithm}
\usepackage{array}
\usepackage[caption=false,position=top,font=normalsize,labelfont=sf,textfont=sf]{subfig}
\usepackage{textcomp}
\usepackage{stfloats}
\usepackage{url}
\usepackage{verbatim}
\usepackage{graphicx}
\usepackage{cite}
\usepackage{times}
\usepackage{epsfig}
\usepackage{amssymb}
\usepackage{makecell}
\usepackage{algpseudocode}
\usepackage{caption}
 \usepackage{subcaption}
\usepackage{times}
\usepackage{amsmath}

\usepackage{xcolor}

\usepackage{tikz}
\usetikzlibrary{trees}

\usepackage{rotating}
\usepackage{lscape}
\usepackage{longtable}   
\usepackage{booktabs}    
\usepackage[breaklinks=true,bookmarks=false]{hyperref}

\begin{document}

\title{Survey on Disaster Management Datasets for Remote Sensing Based Emergency Applications}

\author{Alain P. Ndigande, Josiah Wiggins, Sedat Ozer
\thanks{Alain P. Ndigande was with the Ozer Lab, Dept. of Computer Science, Ozyegin University, Istanbul, Turkiye.}
\thanks{Josiah Wiggins is with the Ozer Lab, Dept. of Electrical and Computer Engineering, California State Polytechnic University, Pomona, CA.}
\thanks{Sedat Ozer is with the Dept. of Electrical and Computer Engineering, California State Polytechnic University, Pomona, CA. Contact email: sedatist@gmail.com \\ This work is accepted for publication at IEEE Transactions on Geoscience \& Remote Sensing.}
}

\maketitle            
\begin{abstract}

Recent natural disasters have highlighted the urgent need for efficient data-driven approaches to disaster management. Machine learning (ML) and deep learning (DL) techniques have shown considerable promise in enhancing the key phases of disaster management including mitigation, preparedness, detection, response, and recovery. A critical enabler of successful ML or DL based applications in remote sensing, however, is the accessibility and quality of annotated datasets. With the growing availability of high-resolution imagery from unmanned aerial vehicles (UAVs) and satellites, computer vision and remote sensing algorithms have become essential tools for rapid detection, situational assessment, and decision-making in disaster scenarios. This survey provides a comprehensive overview of publicly available image-based datasets relevant to ML/DL-based disaster management pipelines. Emphasis is placed on datasets that support computer vision and remote sensing tasks across all phases of disaster events including pre-disaster, during, and post-disaster. The goal of this work is to serve as a centralized reference for researchers and practitioners seeking high-quality datasets for rapid development and deployment of remote sensing-driven disaster response solutions.

\end{abstract}


\section{Introduction}
Natural disasters affect millions of people worldwide, resulting in loss of human life, and in severe damage to infrastructure, property, consequently the economy. On a larger scale, disasters affect not only a few individuals, but also local communities, surrounding cities, and even surrounding countries \cite{guha2015estimating}. Recently, with the increasing occurrences of disasters such as wildfires, tsunamis, earthquakes, hurricanes, tornadoes, droughts, extreme temperature changes, and floods, disaster management and preparedness become vitally important in the modern world.

Disaster management involves pre-, during-, and post-disaster efforts to predict disasters in a timely manner, prevent casualties, and protect both infrastructure and human life. In addition, effective disaster management helps minimize the economic impact and restore stability once hit by a disaster. Given the intricate and complex nature of disasters, robust and informed decision-making is a challenging and important task. Recent technological developments in deep learning (DL) and sensing technologies, can improve effectiveness and accuracy in various mission-critical and decision-making processes.

\begin{figure}[!t]
\begin{center}
 \includegraphics[width=.8\linewidth]{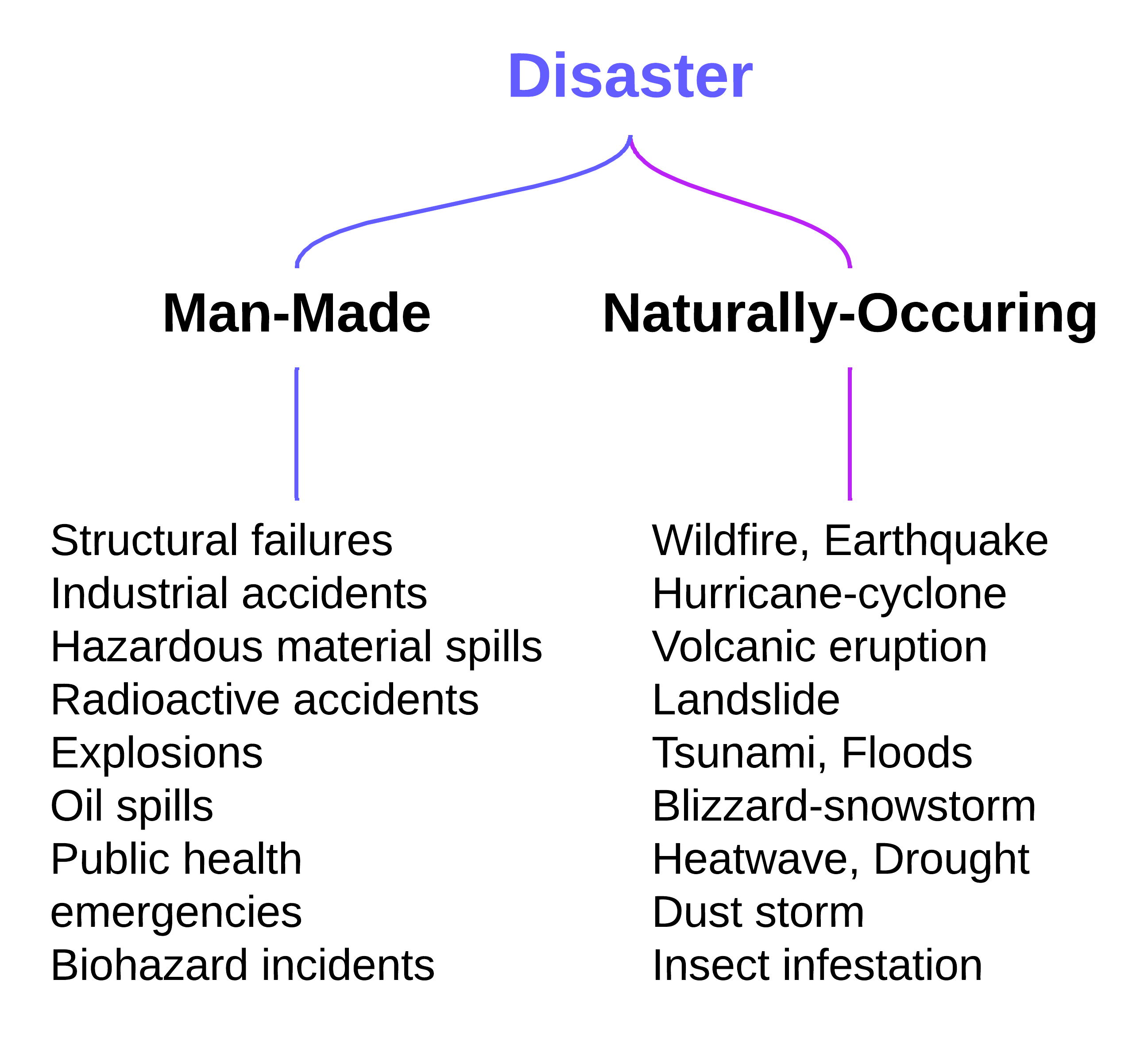}
\end{center}
\vspace{-2mm}
   \caption{A disaster can either be a result of human-related action (man-made) or due to natural causes. This figure provides examples for both of those disaster types. }
\label{fig:DisasterClasses} 
\end{figure}

According to the United Nations Office for Disaster Risk Reduction (UNDRR\footnote{https://undrr.org/terminology/disaster}), a disaster, in summary, is a serious disruption impacting the functioning of a community or a society due to hazardous events, resulting in human, economic, material, or environmental losses. Disasters have been broadly classified as natural or man-made disasters, as depicted in \emph{Figure \ref{fig:DisasterClasses}} and each category may require different types of management and response actions \cite{severin2020types}.

\begin{figure}[!t]
\begin{center}

\includegraphics[width=1.0\linewidth]{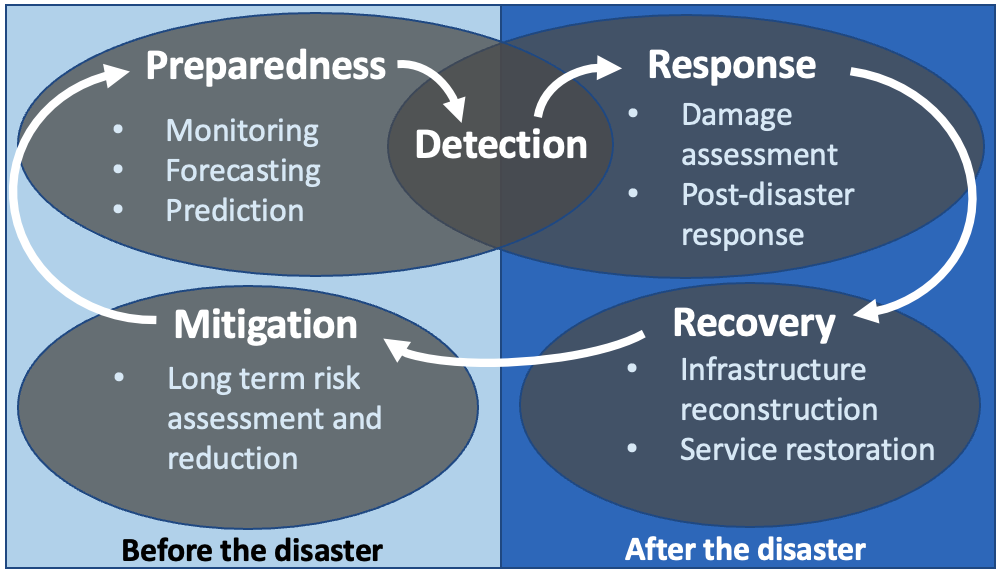} 

\end{center}
\vspace{-2mm}
   \caption{This figure illustrates the five phases of disaster management: mitigation/prevention, preparedness, detection, response, and recovery. These phases form a continuous cycle of planning, detecting, responding to, and recovering from disasters as shown in the figure.}
\label{fig:DisasterManagementCycle} 
\end{figure}

Disaster management involves four main key phases; {\it Mitigation}, {\it Preparedness}, {\it Response}, and {\it Recovery} \cite{sun2020applications, national2007improving, mcentire2021disaster}. The {\it mitigation} phase is a pre-disaster phase and aims to reduce the risk and severity of the disaster(s). It involves actions to be taken beforehand to prevent or minimize the impact of potential disasters, such as implementing building codes, developing preventive laws or regulations, and creating emergency plans \cite{sun2020applications}. The {\it preparedness} phase involves planning, training, and organizing resources to respond effectively when a disaster is likely to occur in a relatively near future. It can include setting up prediction  or detection systems, as well as  activities including drills, the establishment of communication protocols, and the stockpiling of the necessary supplies \cite{sun2020applications}. The {\it response} phase occurs immediately after detecting a disaster,  leading to the deployment of emergency services to provide immediate assistance, evacuations, and implementation of plans to address the immediate needs of affected individuals or communities \cite{sun2020applications}. The {\it recovery} phase aims to restore an affected area to its pre-disaster state (or better), after the disaster. This involves tasks such as rebuilding infrastructure, providing long- or short-term assistance to affected individuals, restoring community services and economic activity \cite{sun2020applications}. In addition to these phases, in many other sources, there is another {\it detection} phase, \cite{lambert2024wildfire}, which typically falls in the overlapping area of preparedness and  response. How these key phases are related to each other is illustrated in Figure \ref{fig:DisasterManagementCycle}. 

In this work, after briefly summarizing different ML and DL techniques used in disaster management, we primarily aim at providing a comprehensive categorization of the available datasets, encompassing their sources, formats, sizes, relevant attributes, availability, and applicability. Furthermore, this survey investigates the challenges associated with dataset collection, dataset availability, annotation, and maintenance for disaster management research. Difficulties such as data heterogeneity, spatio-temporal variations, and imbalances are also discussed, emphasizing the necessity of data pre-processing and augmentation techniques to mitigate these challenges. We aim to explore how designing DL-based applications and datasets connect in handling natural disasters, and as such, this resource will help many researchers and decision-makers better envision the big picture of the available datasets, their usage and provide insights into the applications of computer vision and remote sensing in disaster management phases.

In unexpected or inadequately prepared disaster situations, remote sensing applications play a critical role, particularly during the detection and response phases. Despite their importance, there are relatively few surveys that comprehensively review the existing datasets in this domain. To address this gap, the objective of this survey is to provide a structured overview of the available datasets that can support the development of disaster management algorithms.

Our main contributions with this work are as follows:
(i) a comprehensive introduction to disaster management concepts tailored for the remote sensing community;
(ii) an overview of representative applications across the various phases of disaster management; and
(iii) a curated \& detailed list and analysis of available datasets that can facilitate the development of rapid, remote sensing–based emergency applications. 

\section{Background}

In the context of disaster management, ML and DL algorithms are commonly employed with large volumes of data collected from various sources such as sensors, satellite imagery, social media feeds and historical records. These algorithms can aid in tasks such as predicting the time or location of occurrence, severity of disasters, identifying vulnerable regions or populations, optimizing resource allocation, and assessing the impact of response strategies. Next, we provide a brief introduction to ML and DL terms.

\noindent\textbf{Machine Learning (ML) Techniques:}
Machine learning (ML) involves the development of algorithms and statistical models that learn and make predictions from data \cite{alpaydin2021machine, goodfellow2016deep} to map given input data to its corresponding output. An ML technique is typically categorized as supervised learning-based, semi-supervised learning-based, unsupervised learning-based, or reinforcement learning-based. In supervised learning-based techniques, the algorithm learns from a labeled dataset called training set, where each input sample comes with its corresponding output (label) \cite{alpaydin2021machine,qiu2016survey}. Learning tasks can be classified as classification-based or regression-based, depending on the expected output type. Linear regression is a typical example of a regression algorithm \cite{montgomery2021introduction} that predicts a continuous output by fitting a model to input-output pairs. {\it Decision Trees} and {\it Random Forest} algorithms \cite{ali2012random, zhang2017assessment} can classify the input (feature values) based on one or many hierarchical tree classifiers. {\it Support Vector Machines} (SVM) and many other relevant kernel-based learning algorithms \cite{vapnik1998statistical, jakkula2006tutorial,ozer2018similarity} are large margin-based discriminative models, typically used for classification or regression problems. Semi-supervised learning algorithms can use the available unlabeled data to improve the learning process, besides the labeled data \cite{seeger2000learning}. Unsupervised learning techniques are typically intended to highlight underlying relevant structures, clusters (or patterns) within the unlabeled data \cite{ghahramani2003unsupervised, bengio2012unsupervised}. They are typically used when output labels are not provided in the available dataset. As a particular algorithm, {\it K-Means} divides the data space into {\it K} clusters based on a predefined similarity measure \cite{kodinariya2013review, sinaga2020unsupervised} and assigns each input sample into those {\it K} clusters. Lastly, {\it Reinforcement Learning} (RL) models learn by interacting with an environment and by receiving feedback from the environment in the form of rewards (or penalties) for the actions taken. RL has shown remarkable success especially in the fields of control, robotics and game playing. Recently, they have also been used in computer vision and disaster response problems as in \cite{tsai2019deep,ozer2021visual}.

\noindent\textbf{Deep Learning (DL):} DL methods specifically use learning models inspired by biological neurons where large numbers of neurons are used together to learn and model a complex problem. A typical DL model requires an {\it architecture} which consists of multiple trainable layers and shows how those layers are connected, highlighting the flow of the computation from input to the output in a graphical form. Each layer has its own trainable parameters and hyper-parameters. Trainable layers may contain fully connected layers, convolutional layers, or more complex blocks consisting of combinations of multiple layers, skip connections, and other operations such as normalization, pooling, concatenation, and activation functions \cite{goodfellow2016deep, lecun2015deep, choi2020introduction, nwankpa2018activation}. Recently, DL models have shown remarkable success in learning intricate patterns and representations in many fields, including remote sensing, wireless communications, autonomous systems and natural language processing  \cite{zhu2017deep, zhang2016deep, ozer2024visirnet,valiente2020connected,young2018recent, yildirim2020deep,ilhan2021offloading, torfi2020natural,ersoy2023ortpiece}. Further applications can be found in survey papers such as \cite{qiu2016survey, pouyanfar2018survey}. 

Typical DL models are categorized as {\it Multilayer Perceptron} (MLP) \cite{firat2009generalized}, {\it Convolutional Neural Networks} (CNN) \cite{wu2017introduction}, {\it Recurrent Neural Networks} (RNN) \cite{sherstinsky2020fundamentals}, {\it Generative Adversarial Networks} (GAN) \cite{aggarwal2021generative}, {\it Long-Short-Term Memory} (LSTM) \cite{hochreiter1997long}, {\it Transformers} \cite{ahmad2024comprehensive, aleissaee2023transformers}, and {\it Diffusion Models} \cite{le2024comprehensive, croitoru2023diffusion}. MLPs are a particular type of DL algorithms that consist of multiple layers of Neurons (perceptrons) (MLP), where each neuron in one layer is connected to each neuron in the next layer (i.e., fully connected). MLPs are frequently used in many tasks involving classification and regression \cite{taud2018multilayer}.
CNNs are another type of DL algorithms and they are mainly used to process spatial and visual data, such as images and videos in a more efficient way \cite{albawi2017understanding, o2015introduction}. AlexNet \cite{NIPS2012_c399862d}, GoogLeNet \cite{szegedy2014going}, R-CNN \cite{girshick2014rich}, U-Net and -like architectures along with Hourglass variants \cite{ronneberger2015u, newell2016stacked}, ResNet, SWIN-transformers are some of the revolutionizing architectures and are mostly used as backbones in various vision-based applications. AlexNet and GoogLeNet are primarily used for object classification, while R-CNN and Mask R-CNN \cite{he2017mask} are designed specifically for object detection and segmentation tasks, respectively. CNN-based models are used in many real-time monitoring systems to detect anomalies or changes in environmental conditions providing early warnings and facilitating rapid response efforts \cite{aqib2018disaster, yu2018big}, classify damaged infrastructure, blocked roads, change detection, semantic segmentation and identify individuals needing help, or assess the extent of natural disasters from satellite imagery, drone footage or surveillance cameras \cite{tran2020damage, vinod2022natural}. As an example, Kamilaris et al. \cite{kamilaris2018disaster} predict the direction and speed of wildfires, helping responders prioritize their efforts and resources effectively. Vision Transformers (ViTs) are a relatively recently proposed attention-based approach for visual recognition tasks. ViTs are a type of transformers extended specifically for vision tasks. ViTs effectively capture long-range dependencies within an image, making them a powerful method in computer vision for complex disaster management tasks \cite{dewangan2022figlib, bazi2021vision, guo2022adaln}. An input image is split into a sequence of fixed-size patches (typically $16\times16$ pixels \cite{dosovitskiy2020image}) which are then linearly embedded into vectors (tokens). The core transformer architecture inherently does not capture positional information; therefore, positional embeddings are added to the input tokens to retain spatial (or sequential) information.

RNN and LSTM architectures are widely used for sequential data forms including natural language processing and time series prediction. They are known to be efficient in capturing dependencies in sequential data \cite{sherstinsky2020fundamentals}. GANs belong to the class of generative models within the field of deep learning, which, essentially, learns to generate new data samples from the same distribution of the training data \cite{gui2021review}. GANs consist of two neural network architectures, namely the generator and the discriminator. Both networks are trained simultaneously in an adversarial manner. The generator, essentially, takes random noise or input and generates synthetic data. The discriminator, on the other hand, is designed to distinguish the real data from the generated (synthetic) data produced by the generator \cite{ goodfellow2014generative, creswell2018generative}. GANs can include CNN blocks in both generator and discriminator networks to generate high-quality synthetic images \cite{zhang2019gcgan, shamsolmoali2021image, alqahtani2021applications}.

In general, deep learning techniques are vital tools in processing various types of data such as satellite imagery \cite{kaur2022review, dotel2020disaster}, aerial footage \cite{kamilaris2018disaster}, textual information \cite{kabir2019deep, bhoi2020deep}, and sensor data by automatically extracting features and patterns from these data sources. The ability of deep learning algorithms to handle high-dimensional data and to learn hierarchical representations makes them invaluable for enhancing situational awareness, decision making, and response planning in disaster scenarios \cite{pouyanfar2019multimodal, alidoost2017application, tran2020damage}. This survey focuses on vision based (UAV and satellite-taken imagery) disaster datasets that can be used by various vision-based deep learning techniques.

Another dimension worth mentioning is ethical considerations in AI-driven disaster management systems. Disaster management has been increasingly studied as an \emph{AI for social good} domain \cite{odubola2025ai, pulivarthy2025bias}. AI-driven disaster management represents a high-impact application domain for advancing Responsible AI principles. Recent literature suggests that remote sensing–based emergency systems must explicitly address fairness, transparency, accountability, privacy, and safety, given their direct societal consequences \cite{abid2025ai, vinuesa2022responsible, floridi2022capai}. Similarly, transparency and explainability are critical for operational trust in disaster-response systems \cite{ghaffarian2023explainable}. There are explainable AI techniques applied to satellite-based damage mapping and early warning systems enable practitioners and decision-makers to interpret model outputs, validate predictions, and support accountable decision-making during emergencies \cite{ghamisi2024responsible, ,samek2021explaining, sakthi2025xai}. Additionally, the use of high-resolution remote sensing and crowd-sourced data raises privacy and security concerns, necessitating responsible data governance, anonymization, and geo-privacy preservation mechanisms \cite{mittelstadt2019principles}.

Besides the applied machine learning and deep learning technique(s), another dimension is the variability and availability in data source for disaster management applications. Table \ref{tab:remotesensingsources} summarizes key satellite-based and geospatial data sources. Access to reliable data sources is critical for effective disaster management models. Major satellite-based providers include \textit{USGS Earth Explorer} and \textit{NASA EarthData}, offering extensive archives of Landsat, MODIS, and ASTER imagery. The \textit{Copernicus Dataspace Ecosystem} serves as a primary access point for Sentinel missions (Sentinel-1, 2, 3, 5P), providing crucial SAR and optical data. For vector data and mapping, \textit{OpenStreetMap} and \textit{Natural Earth Data} offer essential baseline geographical information. Commercial providers, such as \textit{Maxar}, supply high-resolution imagery often used for detailed damage assessment. These datasets are typically available in standard geospatial formats, such as GeoTIFF for raster data and Shapefile or GeoJSON for vector data, facilitating integration into machine learning and deep learning pipelines.

\begin{figure}[!t]
\begin{center}
\includegraphics[width=0.9\linewidth]{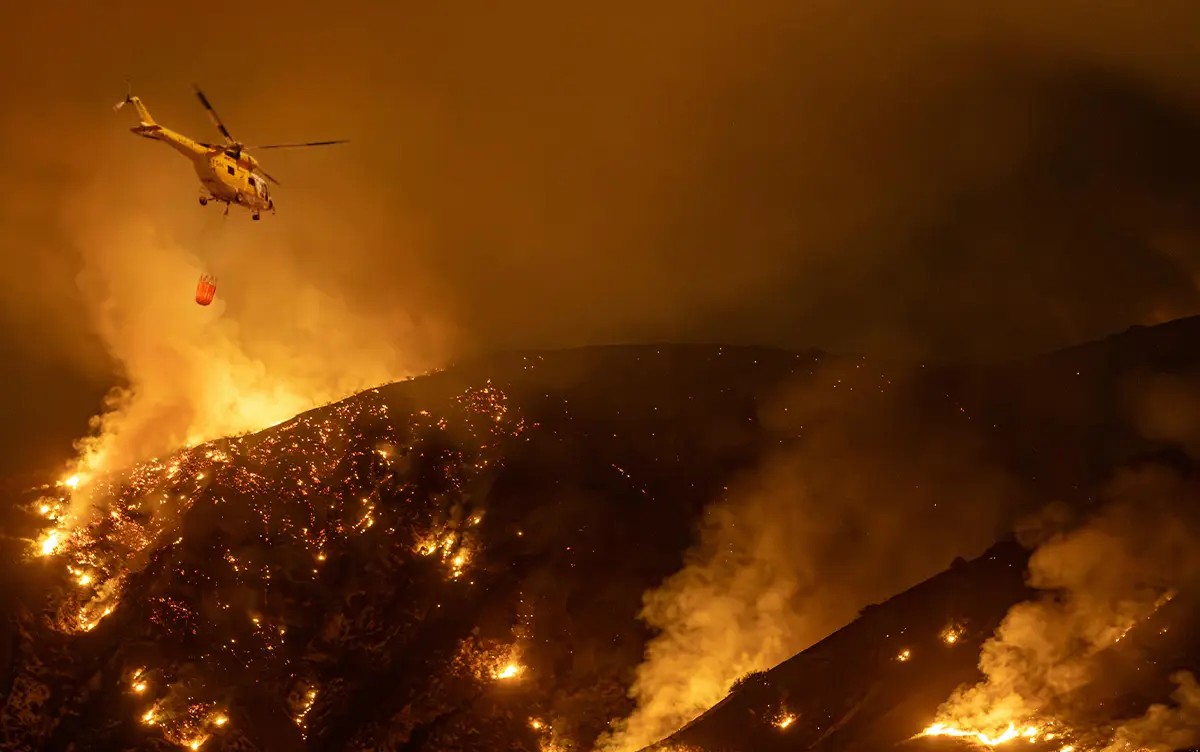} 
\end{center}
   \caption{NOAA emergency response aerial imagery of fire in Los Angeles California fires (Jan 2025) from NOAA’s emergency response flights \cite{noaa_california_2025_fires}.}
\label{fig:california-fires} 
\vspace{-6mm}
\end{figure}


\section{Related work} 
The integration of machine learning (ML) and deep learning (DL) into disaster management has emerged as a powerful paradigm for enhancing decision-making across the disaster lifecycle. While prior surveys have explored other modalities such as textual data \cite{alam2021humaid, kabir2019deep}, this work focuses on remote sensing and computer vision approaches utilizing UAV and satellite imagery.

Benchmark datasets have been instrumental in advancing these capabilities. For instance, the {\it SpaceNet\footnote{https://spacenet.ai/}} challenge series provides open datasets for tasks ranging from building segmentation \cite{van2018spacenet} and road extraction \cite{spacenet5} to multi-temporal urban change detection \cite{Van_Etten_2021_CVPR} and flood detection \cite{hansch2022spacenet}. These resources drive innovation in \textit{disaster mitigation, preparedness, and response}, as well as urban planning.

We organize the literature by disaster management phase: mitigation and preparedness, response, and recovery. Within each phase, works are grouped by disaster type (e.g., floods, wildfires, earthquakes) to highlight methodological patterns.

\subsection{Mitigation and preparedness}
Mitigation and preparedness are critical for reducing disaster risk and ensuring readiness before an event occurs. In this domain, deep learning techniques analyze visual data to identify vulnerabilities in infrastructure and predict potential risks \cite{tan2021can}. These models facilitate the creation of predictive frameworks that estimate impact, thereby aiding in strategic resource allocation and planning. Key applications include early detection systems, hazard mapping, and the development of risk-aware inventories.

\noindent\textbf{Wildfire detection and monitoring:} Early warning systems are critical for minimizing damage. Traditional sensor-based methods (thermal, smoke) are often sparsely distributed or prone to failure \cite{jadon2019firenet}, making camera-based techniques a robust alternative \cite{barz2019enhancing}. For example, \cite{muhammad2018early} introduced a CNN-based framework for fire detection using CCTV surveillance. Plaza et al. \cite{PAOLETTI2019279} proposed a hybrid CNN-RNN framework for wildfire monitoring using hyperspectral data. More recently, Dewangan et al. \cite{dewangan2022figlib} introduced \textit{SmokeyNet}, utilizing CNNs, LSTMs, and Vision Transformers to detect smoke in image sequences, supporting both early warning and continuous monitoring.

\noindent\textbf{Drought and Landslide monitoring:} Deep learning also aids in monitoring slow-onset disasters. Shen et al. \cite{SHEN201948} developed a model to monitor drought regions by integrating multi-source remote sensing data (MODIS \cite{MODIS}, TRMM \cite{TheTropicalRainfallMeasuringMissionTRMMSensorPackage}), capturing both meteorological and agricultural indicators. For landslides, DL methods characterize historical occurrences to map future risk. Carri\'on-Mero et al. \cite{ijerph18189445} utilized CNNs for susceptibility mapping, while Mo et al. \cite{mo2023lightweight} proposed LP-YOLO, a lightweight algorithm for detecting landslide-prone areas using historical Google Earth imagery.

\noindent\textbf{Flood risk and Infrastructure mapping:} Accurate pre-event mapping is essential for risk modeling. H{"a}nsch et al. \cite{hansch2022spacenet} introduced \textit{SpaceNet 8}, combining building footprint and road extraction with flood detection to support infrastructure resilience. Similarly, \textit{FloodNet} \cite{rahnemoonfar2020floodnet} provides high-resolution annotated UAS imagery for segmenting flood-prone areas. Tan et al. \cite{TAN2024113956} built on these to propose a framework for complex urban flood mapping. Furthermore, initiatives like xView2\footnote{https://xview2.org/} \cite{gupta2019xbd} generate large-scale building inventories. While often used for post-disaster assessment, these pre-event baselines are vital for exposure assessment and emergency planning.

\subsection{Response}
The response phase prioritizes situational awareness, including rapid detection, victim localization, and accessibility analysis.

\noindent\textbf{Real-time flood detection:} Identifying inundated regions in near real-time is crucial for response. Munawar et al. \cite{munawar2021uavs} demonstrated the use of UAVs equipped with CNNs for flood detection, achieving 91\% accuracy. While optical data offers high resolution, it is limited by cloud cover. To address this, Rambour et al. \cite{rambour2020flood} proposed a dataset combining optical and Synthetic Aperture Radar (SAR) imagery, enabling robust detection even in adverse weather conditions.

\noindent\textbf{Victim and object localization:} Rapidly locating survivors and critical objects is essential within the critical 72-hour window \cite{toby2022survey}. Chaudhuri et al. \cite{chaudhuri2020exploring} utilized deep learning on geo-tagged images from smart infrastructure to identify survivors in debris. Advanced spectral methods have also been applied; for instance, \cite{Shanjun} introduced a framework merging spatial and spectral features via stacked autoencoders \cite{10.5555/1756006.1953039} and CNNs \cite{rs15184554} to classify objects of interest in complex disaster scenes.

\noindent\textbf{Logistics and Accessibility:} Assessing road networks is vital for routing rescue missions. The pre- and post-event imagery from datasets like xBD \cite{gupta2019xbd} allows for the rapid extraction of road networks and flood extents, directly supporting logistics and on-the-ground operations, as seen in responses to recent major earthquakes.

\subsection{Recovery and damage assessment}
Recovery involves detailed damage assessment and reconstruction planning, often relying on change detection between pre- and post-event imagery.

\noindent\textbf{Damage assessment (Earthquakes, Hurricanes, Floods):} Automated assessment helps prioritize resources based on severity \cite{lozano2023data}. Yanbing et al. \cite{Yanbing} utilized post-event dual-polarimetric SAR for building damage assessment following the 2015 Nepal earthquake. Similarly, Wang et al. \cite{s23146342} proposed a two-stage approach for the 2023 Turkiye earthquake sequence, performing detection and damage classification on optical satellite images. Broader multi-hazard approaches include Msnet \cite{zhu2021msnet}, a multi-level instance segmentation network designed to assess damage from earthquakes, floods, and fires using aerial video.

\noindent\textbf{Scenario analysis and Future planning:} Beyond immediate assessment, deep learning supports long-term impact analysis. Schmidt et al. \cite{schmidt2022climategan} proposed \textit{ClimateGAN} to simulate extreme floods on real-world images, enabling "what-if" visual analysis for urban planning. Additionally, safety systems like the earthquake hazard recognition model by Amin et al. \cite{amin2021earthquake} inform both retrofitting (mitigation) and post-event safety inspections.

Above-mentioned sample studies highlight the critical role of remote sensing and deep learning across the disaster cycle. The following sections provide a detailed categorization of relevant datasets, clustering them by modality and application to guide future research.

\begin{figure}[!t]
\begin{center}
\includegraphics[width=0.9\linewidth]{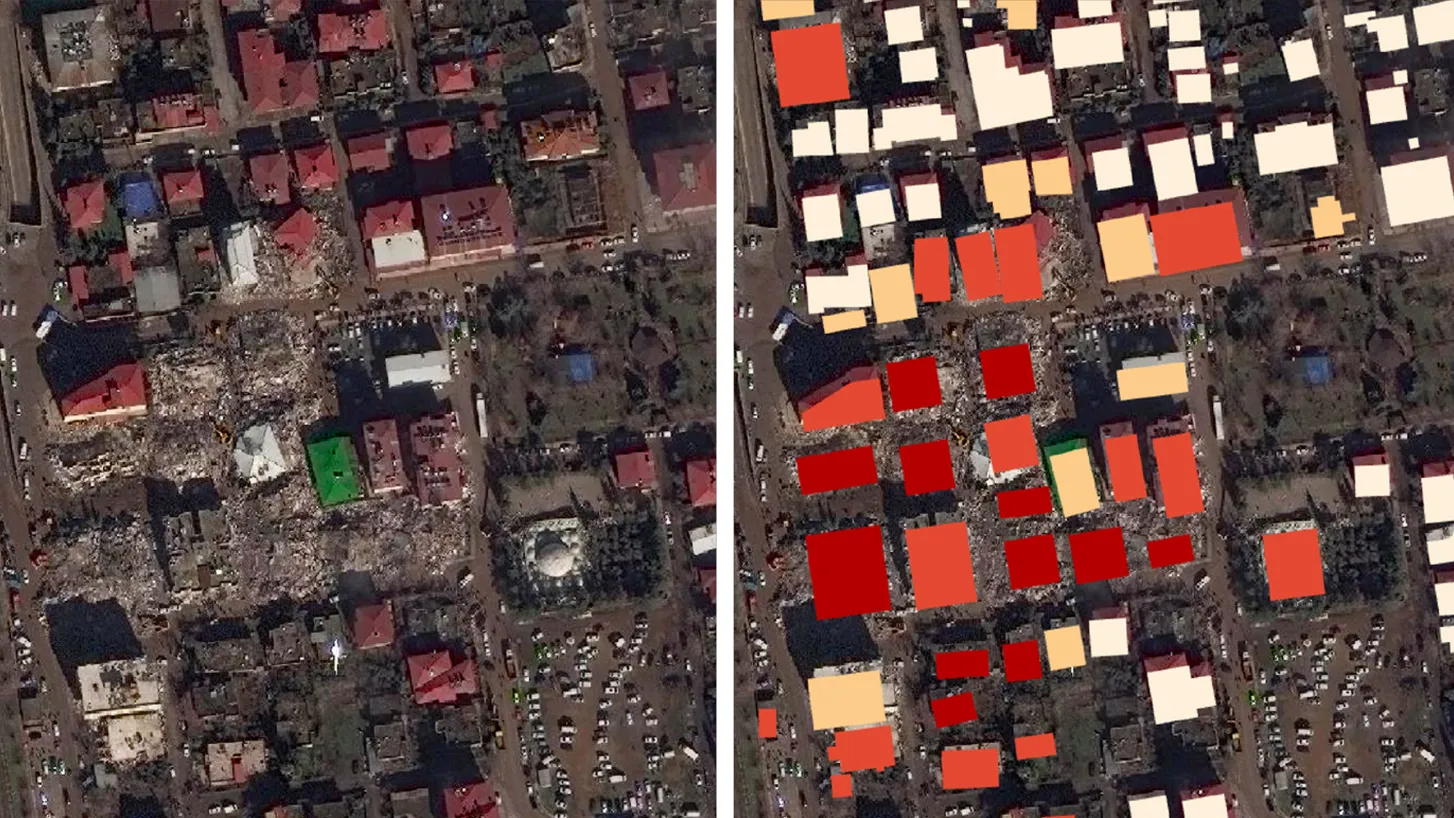} 
\end{center}
\vspace{-2mm}
   \caption{A sample Maxar satellite imagery (left) and its segmented version with the output from xView2 (right) for the earthquake in Islahiye, Turkiye.                  
   }
\label{fig:satelliteImagery} 
\end{figure}

\begin{table*}[ht!]
      \centering
       \caption{This table summarizes satellite-based sources and the data formats for the listed disaster datasets in Table \ref{tabl:all_datasets_together}.}
      \begin{minipage}[b]{0.96\textwidth}
        \setlength\tabcolsep{.92em}
        \begin{tabular}{|>{\centering}p{2.8cm}|m{10.6cm}|p{2.8cm}|}
        \hline
            \textbf{Source} & \textbf{Description} & \textbf{Formats} \\[5pt] \hline 
            \href{https://www.naturalearthdata.com/}{Natural Earth Data}  
            & A public data portal offering both vector and raster datasets for physical and cultural geography. It includes a variety of geospatial data, such as boundaries, coastlines, rivers, lakes, and land cover in varying resolutions (1:10m, 1:50m, and 1:110m). 
            & \begin{tabular}{c} GeoJSON, GeoTIFF \end{tabular} \\ \hline
            
            \href{https://earthexplorer.usgs.gov/}{USGS Earth Explorer} 
            & provides access to a wide range of satellite and aerial imagery datasets, including {\it Landsat}, {\it Sentinel}, {\it MODIS} and {\it ASTER} missions and aerial imagery datasets. USGS hosts time-series data, allowing users to analyze changes in land cover, vegetation, and other variables. 
            & \begin{tabular}{c} GeoTIFF, JPEG, \\KMZ/KML, GeoJSON \end{tabular} \\ \hline 
            
            \href{https://www.openstreetmap.org/}{OpenStreetMap} 
            &  publicly available and crowd-sourced maps with highly detailed free GIS data with different levels of accuracy and completeness. Covers high-spatial-resolution data for buildings, roads, and waterways.  
            & \begin{tabular}{c}OSM XML, PBF, \\Shapefile \end{tabular}\\ \hline
            
            \href{https://www.esri.com/en-us/home}{Esri Open Data Hub} 
            &  A commercial GIS organization providing access to a range of geospatial dataset resources in vector and raster formats. APIs provided to access the data are OGC WMS, GeoJSON, and GeoService.  & \begin{tabular}{c}GeoJSON, Shapefile, \\ CSV, KML \end{tabular} \\ \hline  
            
            \href{https://www.earthdata.nasa.gov/centers/sedac-daac}{NASA’s SEDAC} 
            & NASA’s Socioeconomic Data and Applications Center (SEDAC), a data center within NASA's Earth Observing System Data and Information System (EOSDIS). SEDAC provides socioeconomic and environmental data, as well as tools and services for their analysis. 
            & \begin{tabular}{c} GeoTIFF, Shapefile,\\NetCDF \end{tabular}\\ \hline
            
            \href{https://unepgrid.ch/}{Environmental Data Explorer} 
            & A portal to access high spatial resolution topographic and environmental datasets and tools. It gives spatial and non-spatial data on population, forests, emissions, disasters, and GDP. 
            &\begin{tabular}{c} CSV, Shapefile,\\GeoTIFF \end{tabular}\\ \hline  
            
            \href{https://dataspace.copernicus.eu/}{Copernicus Dataspace Ecosystem} 
            & Successor to the Sentinel Open Access Hub, providing free access to the full archive of Copernicus Sentinel missions (Sentinel-1, -2, -3, -5P) and Copernicus Services. It offers cloud-based processing tools and APIs to facilitate rapid access to Earth observation data for disaster monitoring.
            & \begin{tabular}{c} SAFE, NetCDF, \\ GeoTIFF, JSON, \\ PNG, JPEG \end{tabular} \\ \hline
            
            \href{https://search.earthdata.nasa.gov/search}{NASA EarthData Search} 
            & A platform provided by NASA that enables search, discover, access, and download a wide range of Earth science data products, including remote sensing data collected by satellites and aircraft. In addition to raw sensor data, the platform offers derived data products. 
            & \begin{tabular}{c}HDF5, GeoTIFF, \\CSV, KML/KMZ \end{tabular}\\ \hline
            
            \href{https://coast.noaa.gov/}{NOAA Data Access Viewer} 
            & The National Oceanic and Atmospheric Administration (NOAA) program that allows exploration, visualization, and download of various geospatial datasets, satellite, aerial and/or LiDAR imagery.
            & \begin{tabular}{c} GeoTIFF, JPEG,\\KMZ/KML \end{tabular}\\ \hline 
            
            \href{https://www.maxar.com/open-data}{Maxar Open Data Program} 
            & Access to high-resolution satellite imagery captured by Maxar's constellation, including data from the WorldView series. Maxar’s Open Data Program provides satellite imagery in case of natural disasters such as wildfires, floods, hurricanes, typhoons, and earthquakes, and more.
            & \begin{tabular}{c} GeoTIFF, JPEG,\\KMZ/KML\end{tabular}\\ \hline
            
            \href{https://worldview.earthdata.nasa.gov/}{NASA WorldView} 
            & Provided by NASA's Earth Observing System Data and Information System (EOSDIS). It allows interactive browsing, visualization, and analysis of satellite imagery and data products from NASA's fleet of Earth observation satellites, offering near-real-time access to satellite imagery.
            & \begin{tabular}{c} GeoTIFF, JPEG,\\KMZ/KML\end{tabular}\\ \hline
        \end{tabular}
        \label{tab:remotesensingsources}
    \end{minipage}%
\end{table*}

\section{Acquisition Types and Data Sources}
Disaster management benefits significantly from diverse data modalities originating from a variety of sources, including high-resolution satellite imagery, weather stations, sensor networks, social media, and other global monitoring systems. These datasets vary in nature and are typically selected based on the specific task and the relevant phase within the disaster management cycle.

\noindent\textbf{Satellite and UAV/Drone Imagery:} Aerial imagery covering large areas can be obtained via high-resolution satellites, drones, and Unmanned Aerial Vehicles (UAVs). Both platforms provide imagery in multi-spectral modalities, which can be stored as still images or video. For example, in January 2025, NOAA’s National Geodetic Survey (NGS) deployed emergency-response aircraft over wildfire-impacted areas in Southern California (see Figure \ref{fig:california-fires}). The resulting high-resolution aerial imagery is available through NOAA’s Emergency Response Imagery portal \cite{noaa_california_2025_fires}.

Remote sensing modalities, such as LiDAR, SAR, and RADAR, are utilized for terrain mapping, change detection, identifying structural damage, and assessing the severity of affected regions. \emph{Table \ref{tab:remotesensingsources}} lists various satellite-based data sources. These platforms provide critical geospatial data and metadata that reveal vital information about the extent of damage during and after disasters, supporting rapid response and recovery efforts.

\noindent\textbf{GIS and Sensor Networks:} Sensor networks collect real-time data from various locations, such as weather stations and seismic monitors. Geographic Information Systems (GIS) \cite{burrough2015principles} provide essential tools for storing, visualizing, analyzing, and interpreting this geographic data. GIS data is typically categorized into vector data, which encodes points, lines, and polygons with associated attributes, and raster data, which represents imagery and continuous fields. Sensor networks offer timely measurements that serve as a powerful tool for building intelligent disaster management systems, particularly for early warning and monitoring.

\noindent\textbf{Social Media and Internet:} Social media platforms provide streaming data that is particularly useful during the disaster response phase for understanding the immediate impact. This primarily consists of textual data obtained via Application Programming Interfaces (APIs) from social media and news outlets. In disaster management, this data is used for public sentiment analysis, identifying immediate needs, and providing real-time updates from affected areas. Additionally, crowd-sourced data including individually contributed photos, videos, and reports through crowd-mapping platforms enhances real-time situational awareness and aids in damage assessment.

The advantage of utilizing technical remote sensing resources, such as satellite, UAV, and GIS data, lies in the presence of standardized metadata, which ensures reliability for data fusion efforts. Structured formats (in \emph{Table \ref{tab:remotesensingsources}}), embed crucial information regarding the geographic location and terrain, enabling interpretation of geo-referenced data. Geospatial metadata is explicitly designed to describe the identification, quality, extent, and spatial and temporal aspects of geographic information. In contrast, social media or the general internet data is often noisy, unstructured, and lacks standardized formats. Although social media platforms generate a massive volume of images, the resulting imagery highly varies quality and is noisy, meaning the volume of truly labeled and relevant data is quite small. This heterogeneity adds notable complexity to spatial or temporal registration required during data fusion.

\noindent\textbf{Historical Records and Reports:} Historical disaster records reveal patterns that help predict and understand the nature of recent events. Initiatives such as NOAA's GOES Fire Detection and Characterization (FDC)\footnote{https://www.goes-r.gov/products/baseline-fire-hot-spot.html}, the MODIS Burned Area product\footnote{https://modis-land.gsfc.nasa.gov/burn.html}, and Monitoring Trends in Burn Severity (MTBS)\footnote{https://www.mtbs.gov/} continuously map and serve high-quality disaster data and metadata, such as burn severity and extent. Furthermore, international organizations and global databases, such as the United Nations (UN) and the World Bank\footnote{https://www.undrr.org/publication/documents-and-publications/demographic-intelligence-disaster-risk-reduction-estimating}\textsuperscript{,}\footnote{https://microdata.worldbank.org/index.php/catalog/454}, provide comprehensive data on disaster occurrences, population density, demographics, infrastructure impacts, economic activities, and response efforts across different regions.

\noindent\textbf{Simulations:} Obtaining sufficient real-world data for modeling certain disaster scenarios can be challenging. For instance, tasks such as building damage assessment or flood mapping often require bi-temporal satellite imagery which may not always be available. Simulated data becomes particularly valuable in these contexts where real-world labeled data is scarce. Ou et al. \cite{ou2023method} demonstrated a method for generating synthetic remote sensing images of disaster zones using natural language descriptions, leveraging pre-trained large language models (GPT-4) for captioning and Stable Diffusion for text-to-image generation.

\section{Discussion and Analysis} 
In this section, we present and discuss our methodology, list the selected datasets and analyze their distributions and categorizations. We particularly aim and intend to highlight the relevance of vision-based datasets to four disaster management phases, heterogeneity in data and diverse sources, a dataset-disaster distribution.  Possible applications for available datasets is presented as task diversification for disaster management. This section is organized as follows; we start with providing our data collection methodology followed by a comprehensive overview of the listed datasets in Table \ref{tabl:all_datasets_together}. After that, in the analysis section, we identify existing limitations, coverage gaps, and challenges across current disaster management datasets, providing insights into open research problems and future directions.

\subsection{Data Collection Methodology}
The objective of this survey is to provide a comprehensive and reproducible overview of publicly available datasets suitable for vision- and remote-sensing–based disaster management and emergency response applications. To this end, a systematic data collection and screening strategy was adopted.


\onecolumn
{
\begin{scriptsize}
\setlength\tabcolsep{.26pt}
\begin{longtable}[c]{|>{\centering\bf\arraybackslash} m{2em}|%
                      >{\centering\arraybackslash} m{10.2em}|%
                      >{\centering\arraybackslash} m{5.4em}|%
                      >{\centering\arraybackslash} m{7.4em}|%
                      >{\centering\arraybackslash} m{6em}|%
                      >{\centering\arraybackslash} m{5.6em}|%
                      >{\centering\arraybackslash} m{5.4em}|%
                      >{\centering\arraybackslash} m{6.4em}|%
                      >{\centering\arraybackslash} m{5.6em}|%
                      >{\centering\arraybackslash} m{7.8em}|%
                      >{\centering\arraybackslash} m{7.2em}|%
                      >{\centering\arraybackslash} m{4.4em}|%
            }%
\caption{
    A full dataset detail table. Each row represents a dataset. The "$\sim$" in split and size columns indicate that they are, to the best of our knowledge. In the resolution column {varying} indicate that the resolutions of the image vary. Spatio-temporal indicates that the dataset varies in both spatial and also cover temporal variable.
} 
\label{tabl:all_datasets_together} \\
\hline
    \textbf{ No.} &
    \textbf{Dataset} &
    \textbf{Stage} &        
    \textbf{Disaster} &
    \textbf{Source} &
    \textbf{Modality}&
    \textbf{\# of Samples } &
    \textbf{Split (\%)} &
    \textbf{Annotation Type} &
    \textbf{Resolution} &
    \textbf{Task} &
    \begin{tabular}{c}
        \textbf{Spatial} \\ \textbf{Temporal} \\ \textbf{(Span)}
    \end{tabular} \\
    \hline
    \endfirsthead
    \multicolumn{12}{l}%
    {\tablename\ \thetable\ -- \textit{\bf Continued from previous page}} \\
    \hline
    \endhead
    \hline \multicolumn{12}{r}{\textit{\bf Continued on next page...}} \\
    \endfoot
    \hline
    \endlastfoot
    \hline 
1 &
  California Fires   \cite{noaa_california_2025_fires} &
  Response &
  Wildire &
  Aerial Imagery &
  optical &
  263 &
  $\sim$N/A &
  labels &
  18829 × 18829 &
  Emergency response &
  post \\ \hline 
2 &
  xBD   \cite{gupta2019creating} &
  Response &
  flood, earthquake,   tornado, tsunami, volcanic eruption, hurricane, wild fire &
  Satellite (Maxar DigitalGlobe) &
  optical&
  22,068 &
  80-10-10 &
  masks, labels &
  1024 × 1024 &
  damage assessment &
  \begin{tabular}{c}pre, post\end{tabular}  \\ \hline
3 &
  Sen1Floods11   \cite{bonafilia2020sen1floods11} &
  Mitigation &
  Flood &
  Satellite (Sentinel-S1/S2) &
  S1 bands1-2; S2 bands 1-12 &
  4,831 &
  60-20-20 &
  masks &
  512×512 &
  flood mapping &
  post \\ \hline
4 &
  Global   Very-High-Resolution Landslide Mapping (GVLM) \cite{zhang2023cross} &
  Preparedness &
  Landslide &
  Google Earth &
  optical &
  17 & 
  N/A&
  masks &
  \begin{tabular}{c} Varying $\sim$ in\\ 10808 × 7424 \\and 1748 × 1748\end{tabular} &
  landslide mapping &
  \begin{tabular}{c} pre, post\end{tabular} \\ \hline
5 &
  CrisisMMD   \cite{alam2018crisismmd} &
  Response &
  earthquakes,   hurricanes, wildfires, floods &
  Twitter &
  optical + Text &
  \begin{tabular}{l} 16058 text\\18082 image\end{tabular}  &
  70-15-15 &
  labels &
  varying &
  image and/or text   classification &
  post \\ \hline
6 &
  AIDERv2 \cite{shianios2024direcnetv2} &
  Response &
  Earthquake, fire, flood, normal &
  Aerial Imagery &
  optical &
  16,723 &
  80-10-10 &
  labels &
  \begin{tabular}{c}  224 x 224 \end{tabular} &
  disaster classification &
    pre, post \\ \hline
7 &
  FireRisk    \cite{shen2023firerisk} &
  Preparedness &
  Wilfire &
  Aerial Imagery (Agriculture Imagery Program-NAIP) &
  optical &
  91 872 &
  $\sim$76-24 &
  labels &
  270 × 270 &
  fire risk assessment &
  pre\\ \hline
8 &
  FLAME    \cite{shamsoshoara2021aerial} &
  Response &
  Fire &
  Aerial Imagery &
  optical + IR (videos) &
  cls 87367, seg  5137 &
  cls $\sim$80-20, seg   $\sim$85-15 &
  masks, labels &
  \begin{tabular}{c}Varying $\sim$ in \\      254 × 254,\\      3480 × 2160\end{tabular} &
  fire detection and   segmentation &
    post \\ \hline
9 &
  Fire and Smoke Dataset   \cite{jadon2019firenet} &
  Response &
  Fire &
  Flickr, Google &
  optical (videos) &
  2425 images for training and 62 videos for testing &
  (1124F, 1301NF), (46   Fv with 19094frm,  16 NFv with 6747frm) &
  labels &
  varying &
  fire and smoke   classification &
  post \\ \hline
10 &
  Mila Simulated Floods   \cite{schmidt2022climategan} &
  Response &
  Flood &
  Unity3D, internet &
  optical, depth (simulated) &
  $\sim$20000 &
   N/A&
  labels, depth maps &
  1200 × 900 &
  flood mapping, segmentation &
  pre, post \\ \hline
11 &
  European Flood 2013   Dataset \cite{barz2019enhancing} &
  Response &
  Flood &
  Internet &
  optical, depth &
  3,710 &
   N/A&
  bbox, labels &
  \begin{tabular}{c}512×512,\\      1280×1280\end{tabular} &
  flood detection and   depth estimation &
  post \\ \hline
12 &
  Twitter Flood   \cite{barz2021finding} &
  Response &
  Flood &
  Twitter &
  optical &
  100,800 &
  $\sim$75-0-25 &
  bbox, labels &
  varying &
  flood, pollution, depth estimation  &
  post \\ \hline
13 &
  COCO Earthquake   \cite{bai2021end} &
  Response &
  Earthquake &
  curated &
  optical &
  2,021 &
   N/A&
  masks, labels &
  \begin{tabular}{c}varying,\\      168×300,\\      4600×3070\end{tabular} &
  structural damage   detection &
  post \\ \hline
14 &
  Burned Area   Delineation from Satellite Imagery \cite{colomba2022dataset} &
  Response &
  Fire &
  Satellite (Sentinel-2 L2A) &
  optical &
  73 &
  957 patches &
  masks, labels &
  \begin{tabular}{c}varying,\\      512 × 512\end{tabular} &
  burned area   delineation and damage severity prediction &
    pre, post \\ \hline
15 &
  \begin{tabular}{c}SAD (South \\ Asia Dataset) \cite{arif2020comparative}\end{tabular} &
  Response &
  fire, flood,   Infrastructural, nature, human damage &
  Internet &
  optical &
  493 &
   N/A&
  labels &
  \begin{tabular}{c}varying,\\      150 × 150\end{tabular} &
  disaster image   classification &
  post \\ \hline
16 &
  Image4Act   \cite{alam2020deep} &
  Response &
  earthquake, fire,   flood, hurricane &
  Twitter &
  optical &
  5558 &
  $\sim$40-15-45 &
  labels &
  varying &
  disaster image   classification and severity prediction &
  post \\ \hline
17 &
  FloodNet Dataset   \cite{rahnemoonfar2021floodnet} &
  Response &
  Flood &
  Aerial Imagery &
  optical (aerial image) &
  2343 &
  60-20-20 &
  masks, labels &
  4000 × 3000 &
  flood detection and   mapping &
  pre \\ \hline
18 &
  SeaDronesSee   \cite{varga2022seadronessee} &
  Response &
  marine /coastal &
  Aerial Imagery &
  optical + IR (aerial   image) &
  6062 images, 447400 frames &
  $\sim$58-14-28 &
  bbox, labels &
  \begin{tabular}{c}varying,\\      3840 × 2160,\\      5456 × 3632\end{tabular} &
  maritime object   detection and tracking &
  post \\ \hline
19 &
  Marine Debris Archive   (MARIDA) \cite{kikaki2022marida} &
  Recovery &
  marine /coastal &
  Satellite (sentinel-2) &
  optical + several   bands &
  1381 &
  $\sim$50-25-25 &
  masks &
   256 x 256 &
  maritine debris   detection &
  post \\ \hline
20 &
  ISBDA Dataset   \cite{zhu2021msnet} &
  Response &
  structural damage &
  Aerial Imagery &
  optical (videos) &
  1,030 &
  80-0-20 &
  bbox, labels &
  varying &
  building damage   detection and  severity prediction &
  post \\ \hline
  21 &
  fMoW   \cite{christie2018functional} &
  Preparedness &
  structural damage &
  OSM &
  optical &
  1M+ &
  $\sim$62.85-10.73-12.77 ,   13.65 &
  bbox, labels &
  \begin{tabular}{c}varying,\\      224 × 224\end{tabular} &
  land cover   classification, building detection, road extraction &
  pre, post \\ \hline
22 &
  Calgary-Flood \cite{he2024efficient} &
  Response &
  Flood &
  Aerial Imagery &
  optical &
   2863&
   N/A&
  masks, labels &
  512 x 512 &
  Urban flood mapping &
  during, post \\ \hline
23 &
  Houston-Flood \cite{he2024efficient} &
  Response &
  Flood &
  Aerial Imagery &
  optical &
   2650&
   N/A&
  masks, labels &
  512 x 512 &
  Urban flood mapping &
  during, post \\ \hline
24 &
  ABCD   \cite{fujita2017damage} &
  Response &
  Tsunami &
  Satellite &
  optical &
  10,777 &
  $\sim$80-20 &
  labels &
  \begin{tabular}{c}varying,\\      160 × 160, \\      120 × 120\end{tabular} &
  post tsunami washed   away building classification &
  post \\ \hline
25 &
  CHF2015   \cite{layek2019detection} &
  Response &
  Flood &
  Twitter &
  optical &
  2,667 &
   N/A&
  labels &
  48 × 32 &
  damage assessment &
  post \\ \hline
26 &
  Foggia’s video dataset   \cite{foggia2015real} &
  Response &
  Fire &
  Internet &
  optical &
  104,563 &
  N/A &
  bbox, labels &
  \begin{tabular}{c}varying,\\      320×240, \\     800×600\end{tabular}&
  fire detection &
  post \\ \hline
27 &
  BoWFire   \cite{chino2015bowfire} &
  Response &
  Fire &
  Flickr &
  optical &
  226 &
   N/A&
  labels &
  varying &
  fire and smoke   detection &
  post \\ \hline
28 &
  building damage dataset   \cite{presa2020assessing} &
  Response &
  hurricane &
  Satellite (NOAA, OSM) &
  optical (aerial image) &
  22,513 &
   N/A&
  labels &
  224 × 224 &
  damage assessment &
  pre, post \\ \hline
29 &
  UAV aerial image   \cite{akshya2019hybrid} &
  Response &
  flood &
  Aerial Imagery &
  optical (aerial image) &
  200 &
   N/A&
  labels &
  high-resolution&
  damage assessment &
  pre, post \\ \hline
30 &
  Aerial image data from flooded areas in the state of Texas \cite{yang2019analysis}&
  Response &
  flood &
  USGS HDDS Explorer &
  optical (aerial image) &
  22,891 &
   N/A&
  labels &
  high-resolution&
  damage assessment &
  pre, post \\ \hline
31 &
  Home-Grown Twitter Dataset  \cite{rizk2019computationally} &
  Response &
  earthquake, flood &
  Internet &
  optical &
  1348 &
   N/A&
  labels &
  256 × 256 &
  damage assessment &
  pre, post \\ \hline
32 &
  SUN   \cite{xiao2010sun} &
  Response &
  Infrastructural &
  Internet &
  optical &
  130,519 &
   N/A&
  labels &
  \begin{tabular}{c} $\sim$ \\ 200 x 200\end{tabular}&
  damage assessment &
  pre\\ \hline
33 &
  SpaceNet 2  Building Detection Dataset   \cite{van2018spacenet} &
  Preparedness &
  Infrastructral &
  Satellite (DigitalGlobe,   WorldView 2) &
  optical + NIR +   several bands &
  24586 &
  60-20-20 &
  masks &
  650 × 650 &
  building segmentation &
  pre, post \\ \hline
34 &
  SpaceNet MVOI   (SpaceNet 4) \cite{weir2019spacenet} &
  Preparedness &
  Infrastructral &
  Satellite (DigitalGlobe,   WorldView 2) &
  optical + NIR +   several bands &
  120,000 &
  80-20 &
  masks &
  900 × 900 &
  building segmentation &
  pre, post \\ \hline
35 &
  Road Network   extraction dataset (SpaceNet 5) \cite{spacenet5} &
  Preparedness &
  Infrastructral &
  Satellite (DigitalGlobe, Vivid   Images) &
  optical &
  $\sim$8570 &
  80-20 &
  masks &
  1024 × 1024 &
  road extraction &
  pre, post \\ \hline
36 &
  SpaceNet 7   \cite{Van_Etten_2021_CVPR} &
  Preparedness &
  structural damage &
  Planet satellite imagery mosaics&
  optical &
  90 cubes &
  60-10-20  cubes &
  masks &
  1024 × 1024 &
  change detection &
  pre, post \\ \hline
37 &
  SpaceNet 8   \cite{hansch2022spacenet} &
  Response &
  flood &
  Satellite (Maxar DigitalGlobe) &
  optical &
  $\sim$32000 buildings &
  N/A &
  masks &
  1300 × 1300 &
  building footprint   detection, road network extraction,    flood detection &
  pre, post \\ \hline
38 &
  DeepGlobe 2018 Road   Extraction Dataset \cite{demir2018deepglobe} &
  Preparedness &
  Infrastructral damage &
  Satellite (DigitalGlobe, Vivid   Images) &
  optical &
  8570 &
  $\sim$73-14-13 &
  masks &
  19,584 × 19,584 &
  road extraction &
  post \\ \hline
39 &
  DeepGlobe 2018 Land   Cover Classification \cite{demir2018deepglobe} &
  Preparedness &
  multiple disasters &
  Satellite (DigitalGlobe, Vivid   Images) &
  optical &
  1146 &
  $\sim$70-15-15 &
  masks, labels &
  2448 × 2448 &
  landcover   classification &
  post \\ \hline
40 &
  Bitemporal Image   Classification dataset \cite{dotel2020disaster} &
  Response &
  flood &
  Satellite (Maxar DigitalGlobe) &
  SAR &
  17621 &
   N/A  &
  masks &
  \begin{tabular}{c} 256 × 256 \\ 1024×1024\end{tabular} &
  flood detection and   mapping &
  pre, post \\ \hline
41 &
  Flood Extent Detection   dataset \cite{gahlot2022curating} &
  Preparedness &
  flood &
  Satellite (Sentinel-1/2) &
  SAR &
  $\sim$30k &
  \begin{tabular}{c}3-1-1 \\ locations \end{tabular} &
  masks &
  256 × 256 &
  detecting flood  extent in open waters &
  post \\ \hline
42 &
  SEN12-FLOOD   \cite{rambour2020flood} &
  Preparedness &
  flood &
  Satellite (Sentinel-1/2) &
  optical + SAR &
  \begin{tabular}{c} 337 \\ sequences\end{tabular} &
   $\sim$80-20&
  masks &
  512 × 512 &
  flood detection &
  pre, post \\ \hline
43 &
  Distant object   localization \cite{lee2022distant} &
  Response &
  Multiple Disasters &
  smartphone &
  optical &
  $\sim$18 &
   N/A&
  masks &
   varying &
  object/people   localization from &
  post \\ \hline
44 &
  DEEPFLOOD   \cite{chaudhary2020water} &
  Response &
  flood &
  Twitter, MyCoast &
  optical &
  8145 &
  80-20 &
  labels &
  512 × 512 &
  flood level prediction &
  post \\ \hline
45 &
  Early Fire Detection Dataset   \cite{muhammad2018early} &
  Preparedness &
  fire &
  curated &
  optical &
  68,457 &
  20-80 &
  labels &
  224 × 224 &
  early fire detection   system &
  post \\ \hline
46 &
  Yuan-Long Zhang   Dataset \cite{zhang2023assessment} &
  Recovery &
  multiple disasters &
  Internet &
  optical &
  430 &
  N/A &
  masks &
  513 × 513 &
  disaster waste   identification &
  post \\ \hline
47 &
  Trioucak Cyclone intensity estimation Devaraj \cite{devaraj2021novel} &
  Preparedness &
  hurricane &
  HURSAT &
  NetCDF &
  $\sim$12000 &
  70-20-10 &
  labels &
  50 × 50 &
  hurricane intensity   prediction &
  post \\ \hline
48 &
  Landslide Detection   Dataset  \cite{mo2023lightweight} &
  Preparedness &
  Landslide &
  Google Earth Service &
  optical &
  6895 &
  80-20 &
  bbox, labels &
  640 × 640 &
  landslide detetion &
  post \\ \hline
49 &
  MEDIC   \cite{alam2023medic} &
  Response &
  earthquake, fire, flood, hurricane, landslide &
  Twitter, Google, Bing, Flickr, Instagram &
  optical &
  71,198 &
  69-9-22 &
  labels &
  varying &
  disaster image   classification &
  post \\ \hline
50 &
  Incident1M   \cite{weber2022incidents1m} &
  Response &
  wildfire, flood,   drought, landslide, earthquake &
  Flickr, Twitter &
  optical &
  1,144,148 &
  90-5-5 &
  labels &
  \begin{tabular}{c}varying,\\ 224 × 224\end{tabular} &
  Incident detection   from social media image &
  post \\ \hline
51 &
  CrisisBench   \cite{alam2020deep} &
  Response &
  Earthquake, fire,   flood, hurricane, landslide, other disaster &
  Twitter, Google,   Flickr &
  optical &
 cls 17511, Info 59717, hum 16769, Ds 34896 &
  70-10-20 &
  labels &
  varying &
  disaster image   classification &
  post \\ \hline
52 &
  Damage  assessment Dataset (DAD)   \cite{nguyen2017damage} &
  Response &
  Earthquake, Typhoon,   Hurricane &
  AIDR,  Google &
  optical &
  25,758 &
  60-20-20 &
  labels &
  224 × 224 &
  damage severity   classification &
  post \\ \hline
53 &
  Damage Identification   Multimodal dataset (DMD) \cite{mouzannar2018damage} &
  Response &
  Fires, Floods, Natural landscape, Infrastructural, Human, Non-damage &
  Instagram &
  optical &
  35785 &
  70-30 &
  labels &
  640 × 640 &
  disaster damage   classification &
  post \\ \hline
54 &
  MediaEval 2019   \cite{andreadis2020flood} &
  Response &
  flood &
  Twitter &
  optical + text &
  7077 &
  $\sim$70-30 &
  labels &
  \begin{tabular}{c}varying,\\      512 × 512\end{tabular} &
  flood severity   estimation &
  post \\ \hline
55 &
  Disaster Image Retrieval from Social Media {[}DIRSM{]} \cite{benjamin2017multimedia} &
  Response &
  flood &
  Flickr, Twitter &
  optical &
  6,600 &
  80-20 &
  labels &
   varying &
  flood classification &
  post \\ \hline
56 &
  Flood-Detection in   Satellite Images Dataset (FDSI) \cite{benjamin2017multimedia} &
  Response &
  flood &
  Planet Satellite &
  optical &
  462 &
   N/A &
  masks &
  320 × 320 × 4 &
  flood segmentation &
  post \\ \hline
57 &
  Tianshui city, Gansu   Province, China Dataset \cite{qi2020automatic} &
  Preparedness &
  Landslide &
  Satellite &
  optical + NIR &
  1443 &
  80-20 &
  masks &
  600 × 600 &
  landslide mapping &
  post \\ \hline
58 &
  Victim Detection   Dataset \cite{tham2021joint} &
  Response &
  fire, hurricane,   flood, earthquake, landslide, other &
  Social Media &
  optical &
  8076 &
  $\sim$74-8-18 &
  bbox &
  varying &
  victim detection &
  post \\ \hline
59 &
  Volan 2019 Dataset   \cite{pi2021detection} &
  Response &
  flood, hurricane,   other &
  Aerial Imagery &
  optical (aerial image) &
  875 &
  60-20-20 &
  masks, labels &
  varying &
  disaster damage detection &
  pre, post \\ \hline
60 &
  RDO-Disaster-Images \cite{asif2021automatic} &
  Response &
  multiple disasters &
  Internet &
  optical &
  3787 &
  60-20-20 &
  bbox, labels &
  416 × 416 &
  disaster detection from media images &
  post \\ \hline
61 &
  UAV pre-post disaster imagery for flood detection \cite{munawar2021uavs} &
  Response &
  flood &
  Aerial Imagery &
  optical &
  51600 &
  $\sim$42-58&
  masks, labels &
  256 × 256 &
  flood detection &
  pre, post \\ \hline
62 &
  Khartoum flood dataset   \cite{saleh2023pdca} &
  Response &
  flood &
  Satellite &
  SAR &
  226 &
  $\sim$64-18-18 &
  masks &
  512 × 512 &
  flood detection &
  pre, post \\ \hline
63 &
  EarthNet2021   \cite{requena2020earthnet2021} &
  Mitigation &
  multiple disasters &
  Satellite &
  optical + NIR +   masks &
  $\sim$32000 data cubes &
   N/A&
  masks &
  \begin{tabular}{c} 80 x 80 \\ 128 x 128 \\ 212 × 212 \end{tabular}&
  extreme weather   forecasting &
     post \\ \hline
   64 &
  Next Day Wildfire   Spread Detection \cite{huot2022next} &
  Preparedness &
  fire &
  Satellite &
  masks &
  18545 fire event &
  80-10-10 &
  masks &
  64 × 64 &
  predicting wildfire   spread &
  pre, post \\ \hline
65 &
  VIDI   \cite{sesver2022vidi} &
  Response &
  multiple disasters &
  Internet &
  optical (videos) &
  4534 video &
  $\sim$82-9-9&
  labels &
  720 x 720 &
  video incident   classification &
    post \\ \hline
66 &
  FIVR-200K   \cite{kordopatis2019fivr} &
  Response &
  multiple disasters &
  Internet &
  optical (videos) &
  $\sim$225960 &
  31-69 &
  labels &
  varying &
  Incident video   retrieval &
    post \\ \hline
67 &
  Fire Ignition Library   (FIgLib) \cite{dewangan2022figlib} &
  Preparedness &
  fire &
  camera &
  optical (videos) &
  24,800 &
  $\sim$53-24-23 &
  bbox, labels &
  \begin{tabular}{c}1536 × 2048,\\ 2048 × 3072\end{tabular} &
  real-time wildfire   smoke detection &
    post \\ \hline
68 &
  Hephaestus   \cite{bountos2022hephaestus} &
  Preparedness &
  volcanic eruption,   earthquake &
  Satellite &
  interferogram &
  216,106 &
  $\sim$92-4-4 &
  masks, labels &
  224 × 224 &
  volcanic stage   classification &
  post \\ \hline
69 &
  Efficient Device-Edge   Inference for Disaster Classification dataset \cite{yang2022efficient} &
  Response &
  cyclone, earthquake,   flood, wildfire &
  Internet &
  optical &
  7124 &
  65.5 - 25 -7.5 &
  labels &
  224 × 224 &
  disaster   classification &
  post \\ \hline
70 &
  e FIO-EP dataset   \cite{dong2023enteromorpha} &
  Preparedness &
  Marine /coastal &
  Satellite &
  several bands &
  1334 &
  $\sim$60-20-20 &
  masks &
  512 × 512 &
  maritime detection &
  post \\ \hline
71 &
  MMFlood dataset   \cite{montello2022mmflood} &
  Response &
  flood &
  Satellite &
  SAR &
  $\sim$8522 tiles &
  $\sim$72-7-21   tiles &
  masks &
  \begin{tabular}{c}varying,\\      531 × 524,\\      1944 × 1944\end{tabular} &
  flood mapping &
  post \\ \hline
72 &
  Fire-flood-flickr (3F)   -emergency dataset \cite{giannakeris2018people} &
  Response &
  fire, flood &
  Social Media &
  optical &
  12423 &
  $\sim$80-20 &
  bbox, labels &
  varying &
  severity level   estimation &
  pre\\ \hline
73 &
  etci2021 Sentinel-1   dataset \cite{etci2021} &
  Response &
  flood &
  Satellite &
  SAR &
  66810 VV\&VH pairs &
  $\sim$60-18-22 &
  masks &
  256 × 256 &
  flood mapping &
  post \\ \hline
74 &
  Disaster-Dataset   \cite{niloy2021novel} &
  Response &
  earthquake,   infrastructural, fire, landslide, drought &
  Social   Media &
  optical &
  13,720 &
  $\sim$98--2 &
  labels &
  224 × 224 &
  disaster   image detection &
  post \\ \hline 
75 &
  TUHO dataset  \cite{amin2021earthquake} &
  Mitigation &
  Earthquake (indoor) &
  Smartphone &
  optical &
  24,000 &
  $\sim$80-20 &
  bbox, labels &
  \begin{tabular}{c} varying, \\ 416  ×  416\end{tabular} &
  risk assessment &
  pre\\ \hline
76 &
  DisasterM3 \cite{wang2025disasterm3} &
  Response &
  multiple disasters&
  Satellite &
  optical + SAR &
  24,000 &
  $\sim$78-22 &
  masks, labels &
  \begin{tabular}{c} high-resolution\end{tabular} &
  damage assessment &
   pre, post \\ \hline
 77 &
  RescueADI \cite{liu2025rescueadi} &
  Response &
  hurricane&
  Aerial Imagery &
  optical &
  4,044 &
  $\sim$92-8 &
  masks, labels &
  \begin{tabular}{c} high-resolution\end{tabular} &
  disaster interpretation &
    post \\ \hline
 78 &
  STURM-Flood \cite{notarangelo2025sturm} &
  Response &
  flood &
  Satellite (Sentinel-1/2) &
  optical + SAR &
  185, 277 &
  80-10-10 &
  masks &
  \begin{tabular}{c} 128 × 128\end{tabular} &
  flood mapping &
   pre, post \\ \hline
  79 &
  Sen2Fire \cite{xu2024sen2fire} &
  Response &
  wildfire &
  Satellite (Sentinel-2/5P) &
  optical + several bands &
  2,466 &
  $\sim$60-20-20 &
  masks &
  \begin{tabular}{c} 512 × 512\end{tabular} &
  damage assessment &
  during \\ \hline
  80 &
  Land8Fire \cite{tran2025land8fire} &
  Response &
  wildfire &
  Satellite (Landsat-8) &
  several bands &
  9,000 &
  N/A &
  masks &
  \begin{tabular}{c} 256 × 256 \end{tabular} &
  fire detection and segmentation &
    pre, post \\ \hline

    81 &
  Kuro Siwo \cite{bountos2023kuro} &
  Response &
  flood &
  Satellite  &
  SAR &
  533,847 &
  N/A &
  masks, labels &
  \begin{tabular}{c} 224 × 224 \end{tabular} &
  flood segmentation &
    pre, post \\ \hline
    82 &
  Landslide4Sense \cite{ghorbanzadeh2022landslide4sense} &
  Response &
  Landslide &
  Satellite (Sentinel-2 + ALOS PALSAR)  &
  S2 bands1–12; AP bands13-14 &
  4,844 &
  78-5-17 &
  masks, labels &
  \begin{tabular}{c} 128 × 128 \end{tabular} &
  landslide detection &
    post \\ \hline
    83 &
  BRIGHT \cite{chen2025bright} &
  Response &
  earthquake, hurricane, cyclone, wildfire, flood, volcanic eruption &
  Satellite &
  optical + SAR &
  4,246 &
  70-10-20 &
  masks, labels &
  \begin{tabular}{c} high-resolution \end{tabular} &
  damage assessment &
    pre, post \\ \hline
    84 &
  PDD \cite{song2023pdd} &
  Response &
  Structural damage, human &
  Aerial Imagery &
  optical &
  879 &
  N/A &
  bbox, labels &
  \begin{tabular}{c} high-resolution \end{tabular} &
  survivor detection &
    pre, post \\ \hline
    85 &
  RescueNet \cite{rahnemoonfar2023rescuenet} &
  Response &
  hurricane &
  Aerial Imagery &
  optical &
  4,494 &
  80-10-10 &
  masks, labels &
  \begin{tabular}{c} 3000 x 4000 \end{tabular} &
  damage assessment &
    post \\ \hline
    86 &
  S2Looking \cite{shen2021s2looking} &
  Recovery &
  Infrastructure &
  Satellite &
  optical &
  5000 image pairs &
  70-10-20 &
  labels &
  \begin{tabular}{c} 1024 × 1024 \end{tabular} &
  building change detection &
    pre, post \\ \hline
    87 &
  HurMic-VQA \cite{sarkar2021vqa} &
  Recovery &
  hurricane &
  Aerial Imagery &
  optical + text &
  3,197; 3 question types per image &
  60-20-20 &
  labels, text &
  \begin{tabular}{c} 4000 × 3000 \end{tabular} &
  disaster damage assessment &
    post \\ \hline
    88 &
  DISC \cite{jeon2021large} &
  Response &
  multiple disasters &
  curated &
  optical &
  300,000 &
  N/A &
  labels, depth maps &
  \begin{tabular}{c} high-resolution \end{tabular} &
  object localization &
    pre, post \\ \hline
    89 &
  Maduo Earthquake Crack Dataset \cite{yu2022earthquake} &
  Response &
  earthquake &
  Aerial Imagery &
  optical &
  51,054 &
  N/A &
  labels &
  \begin{tabular}{c} high-resolution \end{tabular} &
  earthquake crack detection &
    post \\ \hline
    90 &
  EBD \cite{wang2025constructing} &
  Response &
  multiple disasters &
  Satellite &
  optical &
  18,215 &
  N/A &
  labels &
  \begin{tabular}{c} 512 x 512 \end{tabular} &
  building damage assessment &
    pre, post \\ \hline
    91 &
  Volan2018 \cite{pi2020convolutional} &
  Recovery &
  hurricane &
  Aerial Imagery &
  optical &
  65,580 &
  N/A &
  bbox, labels &
  \begin{tabular}{c} varying \end{tabular} &
  object detection for disaster recovery &
    post \\ \hline
    92 &
  UNFSI \cite{tang2023survey} &
  Preparedness &
  Infrastructural &
  Aerial Imagery &
  optical &
  5,705 &
  N/A &
  bbox, labels &
  \begin{tabular}{c} 4000 × 2250,\\ 640 × 640 \end{tabular} &
  road crack detection &
    during \\ \hline
    93 &
  Kerala 2018 Monsoon Landslide Inventory Dataset \cite{hao2020constructing} &
  Mitigation &
  landslide &
  Google Earth &
  optical &
  4,728 &
  N/A &
  labels &
  \begin{tabular}{c} high-resolution \end{tabular} &
  landslide inventory mapping &
    post \\ \hline
    94 &
  Bijie landslide dataset \cite{ji2020landslide} &
  Recovery &
  landslide &
  Satelite &
  optical &
  770 &
  67-0-33 &
  masks, labels &
  \begin{tabular}{c} high-resolution \end{tabular} &
  landslide inventory mapping &
    post \\ \hline
    95 &
  Ida-BD \cite{lee2022ida} &
  Recovery &
  hurricane &
  Satellite (WorldView-2) &
  optical &
  89 image pairs &
  N/A &
  labels &
  \begin{tabular}{c}  1024 x 1024 \end{tabular} &
    building damage assessment &
    post \\ \hline
    96 &
  3DAeroRelief Dataset \cite{le20253daerorelief} &
  Recovery &
  hurricane &
  Aerial Imagery &
  optical &
  64 point clouds &
  88-0-12 &
  labels &
  \begin{tabular}{c}  1920 × 1080 \end{tabular} &
  3D damage semantic segmentation &
    post \\ \hline
    97 &
  DisasterScope \cite{liu2024disasterscope} &
  Recovery &
  multiple disasters &
  Aerial Imagery &
  optical &
  1,030 &
  70-20-10 &
  bbox, labels &
  \begin{tabular}{c}  640 x 640 \end{tabular} &
  disaster-related object detection &
    during, post \\ \hline
    98 &
  DoriaNET \cite{cheng2021dorianet} &
  Recovery &
  hurricane &
  Aerial Imagery &
  optical &
  2,409 buildings &
  N/A &
  masks, labels &
  \begin{tabular}{c}  1280 × 720 \end{tabular} &
  building damage assessment &
    post \\ \hline
    99 &
  BanglaCalamityMMD \cite{faria2025banglacalamitymmd} &
  Response &
  Landslides, Wildfire, Tropical Storm, Drought, Flood, Earthquake, Human Damage, Non-Disaster &
  Internet &
  optical + text &
  7,903 &
  80-10-10 &
  labels &
  \begin{tabular}{c}  varying \end{tabular} &
  multi-modal disaster identification &
    during, post \\ \hline
    100 &
  FloodNet-VQA \cite{sarkar2023sam} &
  Recovery &
  flood &
  Aerial Imagery &
  optical &
  2,188 &
  60-20-20 &
  labels &
  \begin{tabular}{c}  224 x 224 \end{tabular} &
  flood damage assessment &
    post \\ \hline
    101 &
  HRUD (High Resolution UAV Dataset) \cite{chowdhury2020comprehensive} &
  Recovery &
  hurricane &
  Aerial Imagery &
  optical &
  1,973 &
  N/A &
  masks, labels &
  \begin{tabular}{c}  3000 × 4000 \end{tabular} &
  segmentation for disaster damage assessment &
    post \\ \hline
    102 &
  RSCC \cite{chen2025rscc} &
  Recovery &
  multiple disasters &
  Satellite &
  optical + text &
  62,351 image pairs &
  N/A &
  labels &
  \begin{tabular}{c}   1024×1024, \\ 512 x 512 \end{tabular} &
  disaster change captioning &
    pre, post \\ \hline
    103 &
  C2A \cite{nihal2024uav} &
  Response &
  traffic incidents, fire, flood, collapsed buildings &
  Aerial Imagery &
  optical &
  10,215 &
  N/A &
  bbox, labels &
  \begin{tabular}{c}   123 × 152 to \\ 5184 × 3456 \end{tabular} &
  human detection in disaster &
    during, post \\ \hline
    104 &
  TSEqD \cite{dar2025social} &
  Response &
  earthquake &
  Social Media &
  optical + text &
  10,352 &
  N/A &
  labels &
  \begin{tabular}{c}   varying \end{tabular} &
  disaster content classification &
    during \\ \hline
    105 &
  MADOS \cite{kikaki2024detecting} &
  Response &
  marine pollution &
  Satellite (Sentinel-2) &
  S2 bands1-13 &
  2,803 &
  N/A &
  masks, labels &
  \begin{tabular}{c}   240 x 240 \end{tabular} &
  marine pollution segmentation &
    during, post \\ \hline
    106 &
  HLS Burn Scars \cite{HLS_Foundation_2023} &
  Recovery &
  wildfire &
  Satellite (Sentinel-2) &
  S2 bands1-6 &
  804 &
  67-0-33 &
  masks, labels &
  \begin{tabular}{c}   512 x 512 \end{tabular} &
  burn scar segmentation &
    post \\ \hline
    107 &
  DODD \cite{zhao2025enhancing} &
  Response &
  multiple disasters &
  curated &
  optical &
  121 &
  N/A &
  bbox, labels &
  \begin{tabular}{c}   varying \end{tabular} &
  human detection under occlusion &
    during \\ \hline
    108 &
  DISASTER dataset \cite{salluri2020object} &
  Response &
  earthquake, flood, cyclone, wildfire &
  curated &
  optical &
  2,423 &
  80-0-20 &
  labels &
  \begin{tabular}{c}   varying \end{tabular} &
  disaster image classification &
    during, post \\ \hline
    109 &
  LADI v2 (Low Altitude Disaster Imagery v2) \cite{scheele2025ladi} &
  Response &
  multiple disasters &
  Aerial Imagery &
  optical &
  9,963 &
  81-9-10 &
  labels &
  \begin{tabular}{c}   varying \end{tabular} &
  multi-label disaster image classification &
    during, post \\ \hline
    110 &
  CRASAR-U-DROIDs \cite{manzini2024crasar} &
  Recovery &
  multiple disasters &
  Aerial Imagery &
  optical &
  21,716 buildings &
  81-0-19 &
  labels &
  \begin{tabular}{c}   high-resolution \end{tabular} &
  building damage assessment &
    post \\ \hline
  
\end{longtable}%
\end{scriptsize}
}
\twocolumn

{
\begin{figure*}[t]
\centering
  \subfloat{%
    \label{fig:source-distr}
    \begin{minipage}[b]{0.192\textwidth}
      \centering
      \includegraphics[width=\textwidth]{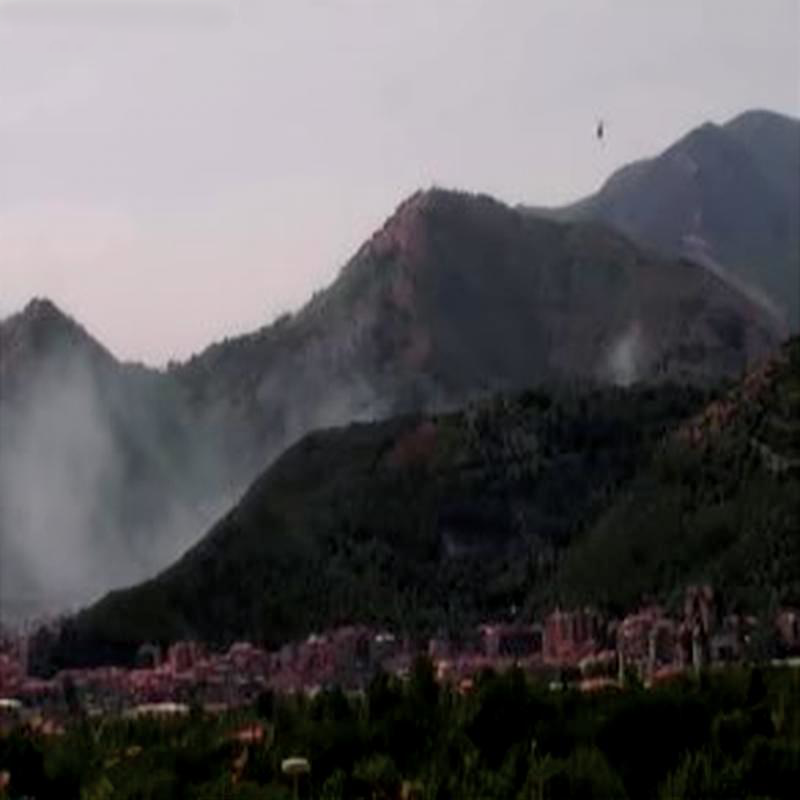}
      \vspace{1mm}
      \parbox{\textwidth}{\centering
      Foggia\cite{foggia2015real}\\[-2pt]
      {\scriptsize fire detection}
      }
    \end{minipage}%
  }
  \!
  \subfloat{%
    \begin{minipage}[b]{0.192\textwidth}
      \centering
      \includegraphics[width=\textwidth]{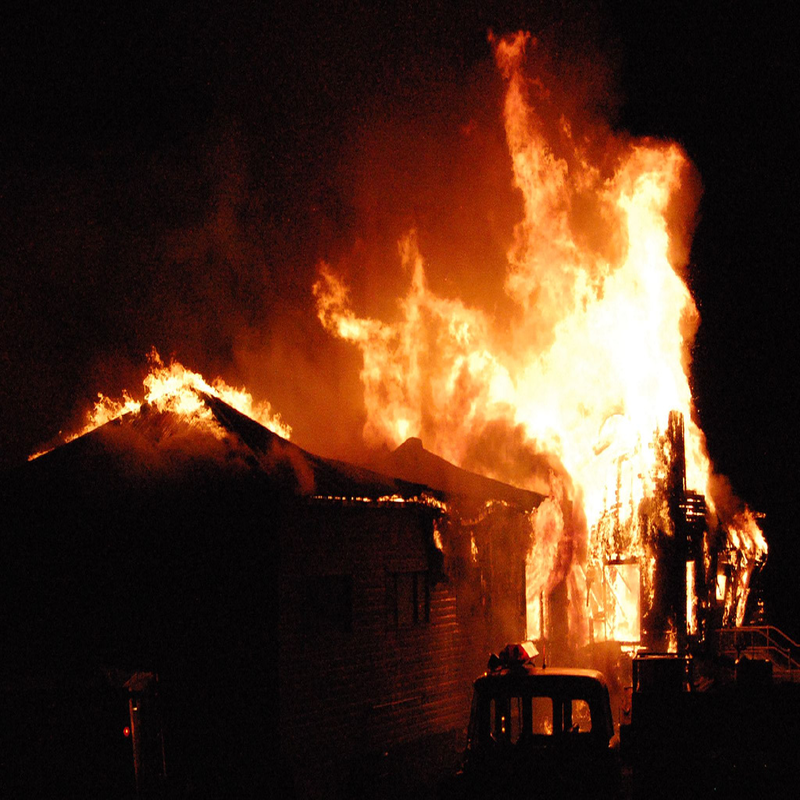}
      \vspace{1mm}
      \parbox{\textwidth}{\centering
      BoWFire\cite{chino2015bowfire}\\[-2pt]
      {\scriptsize fire and smoke detection}
      }
    \end{minipage}%
  }
  \!
  \subfloat{%
    \begin{minipage}[b]{0.192\textwidth}
      \centering
      \includegraphics[width=\textwidth]{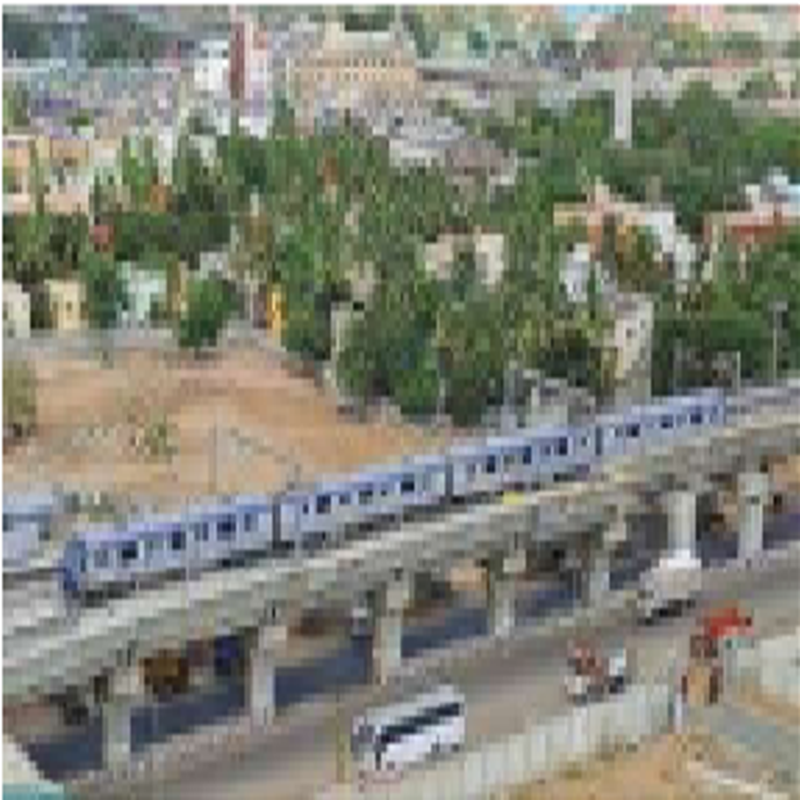}
      \vspace{1mm}
      \parbox{\textwidth}{\centering
      UAV AIms\cite{akshya2019hybrid}\\[-2pt]
      {\scriptsize flood damage assessment}
      }
    \end{minipage}%
  }
  \!
  \subfloat{%
    \begin{minipage}[b]{0.192\textwidth}
      \centering
      \includegraphics[width=\textwidth]{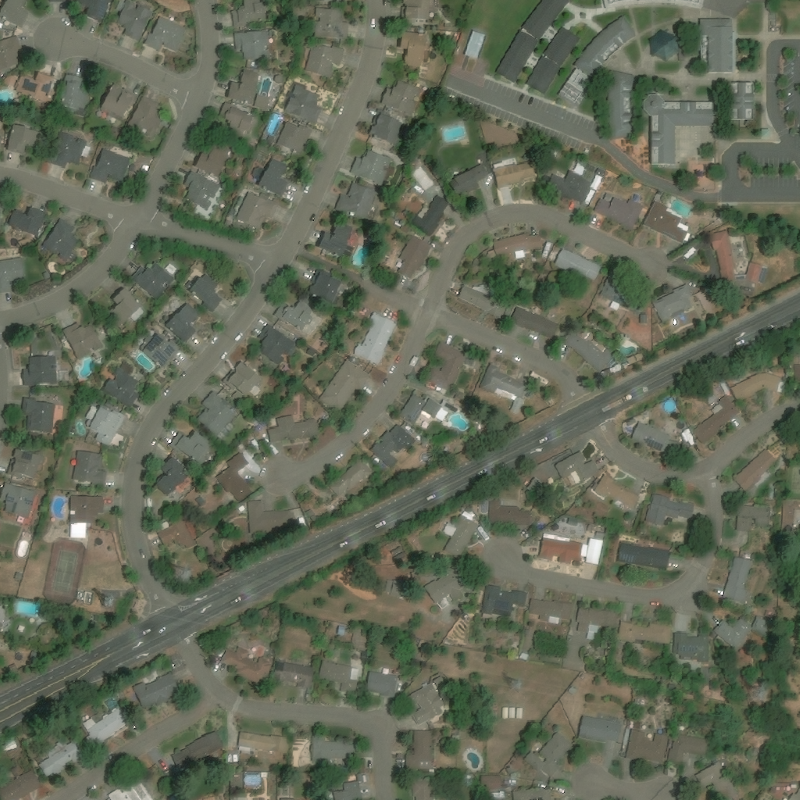}
      \vspace{1mm}
      \parbox{\textwidth}{\centering
      xBD pre\cite{gupta2019creating}\\[-2pt]
      {\scriptsize building damage assessment}
      }
    \end{minipage}%
  }
  \!
  \subfloat{%
    \begin{minipage}[b]{0.192\textwidth}
      \centering
      \includegraphics[width=\textwidth]{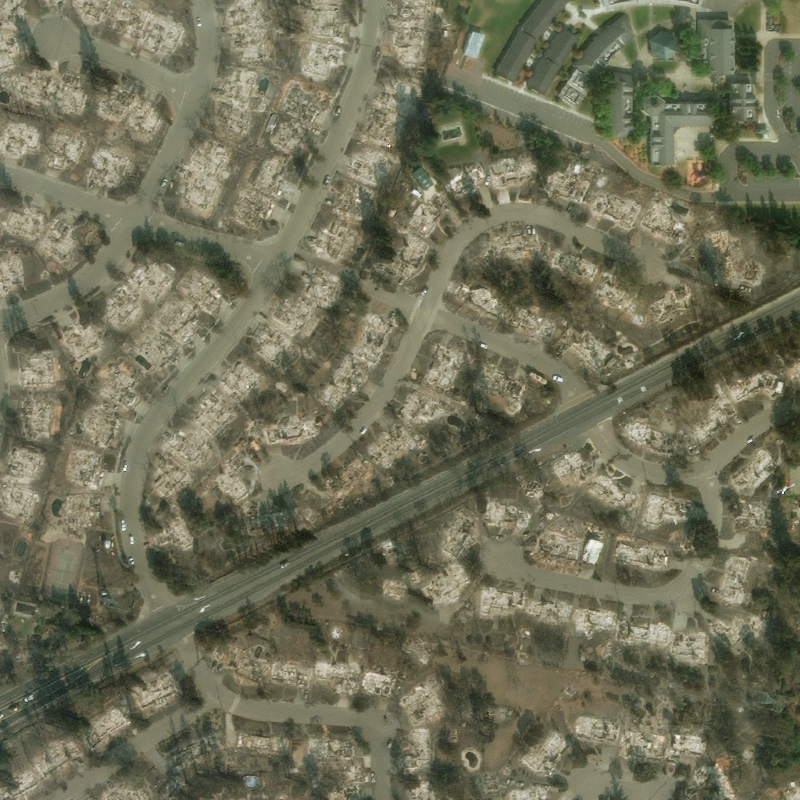}
      \vspace{1mm}
      \parbox{\textwidth}{\centering
      xBD post\cite{gupta2019creating}\\[-2pt]
      {\scriptsize building damage assessment}
      }
    \end{minipage}%
  }

\vspace{-4mm}
  \subfloat{%
    \begin{minipage}[b]{0.192\textwidth}
      \centering
      \includegraphics[width=\textwidth]{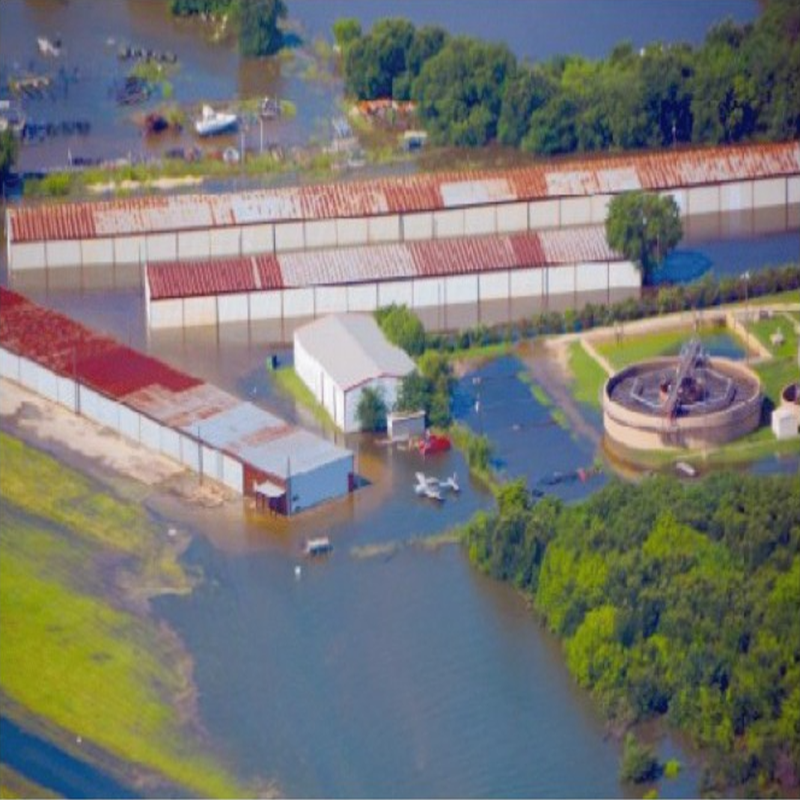}
      \vspace{1mm}
      \parbox{\textwidth}{\centering
      Texas Flood\cite{yang2019analysis}\\[-2pt]
      {\scriptsize flood damage assessment}
      }
    \end{minipage}%
  }
  \!
  \subfloat{%
    \begin{minipage}[b]{0.192\textwidth}
      \centering
      \includegraphics[width=\textwidth]{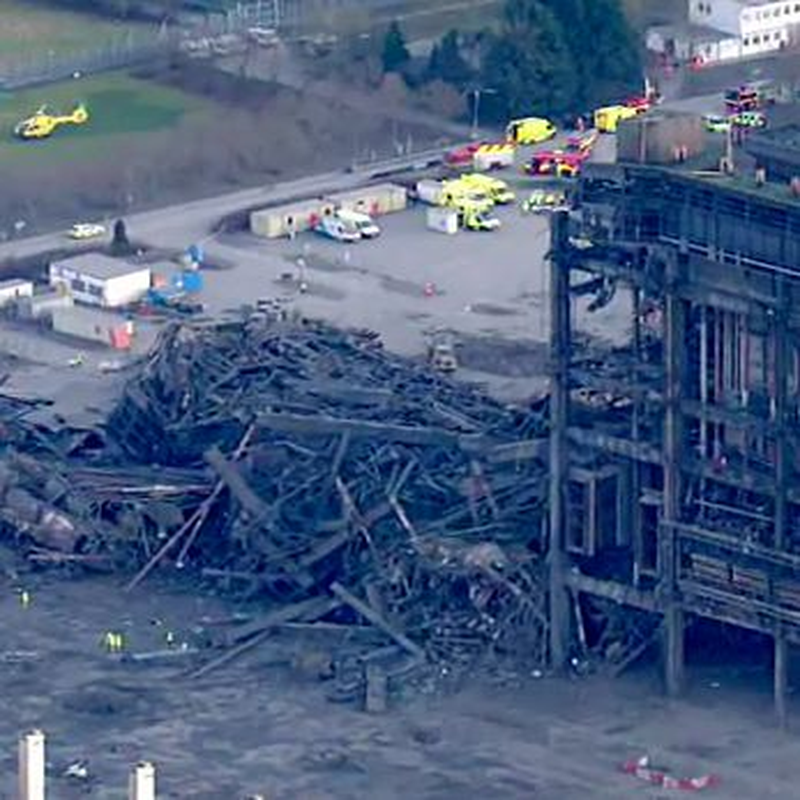}
      \vspace{1mm}
      \parbox{\textwidth}{\centering
      AIDERv2\cite{shianios2024direcnetv2}\\[-2pt]
      {\scriptsize disaster classification}
      }
    \end{minipage}%
  }
  \!
  \subfloat{%
    \begin{minipage}[b]{0.192\textwidth}
      \centering
      \includegraphics[width=\textwidth]{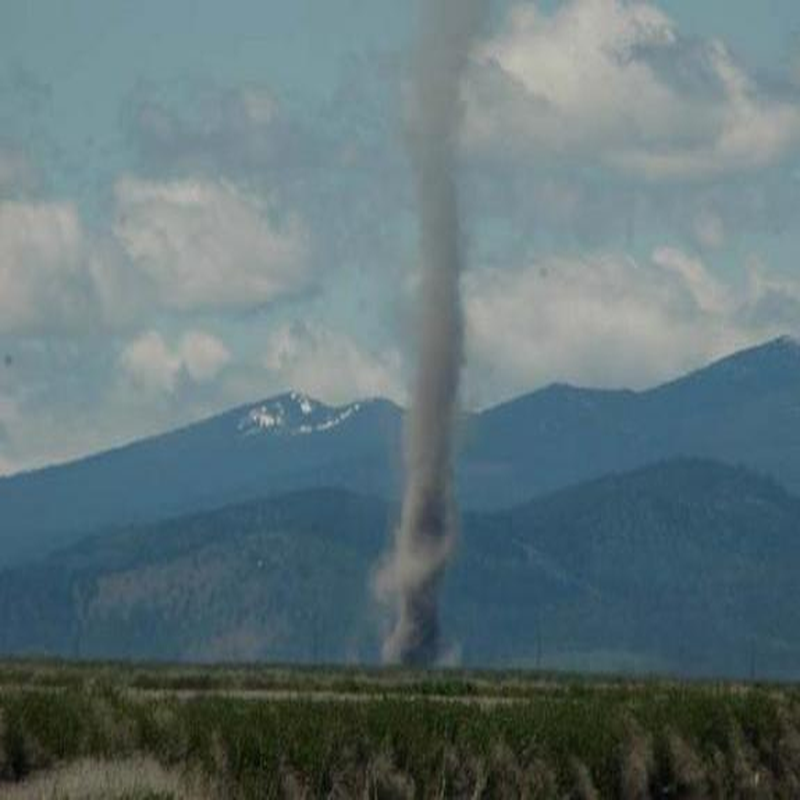}
      \vspace{1mm}
      \parbox{\textwidth}{\centering
      Incidents\cite{weber2020detecting}\\[-2pt]
      {\scriptsize disaster classification}
      }
    \end{minipage}%
  }
  \!
  \subfloat{%
    \begin{minipage}[b]{0.192\textwidth}
      \centering
      \includegraphics[width=\textwidth]{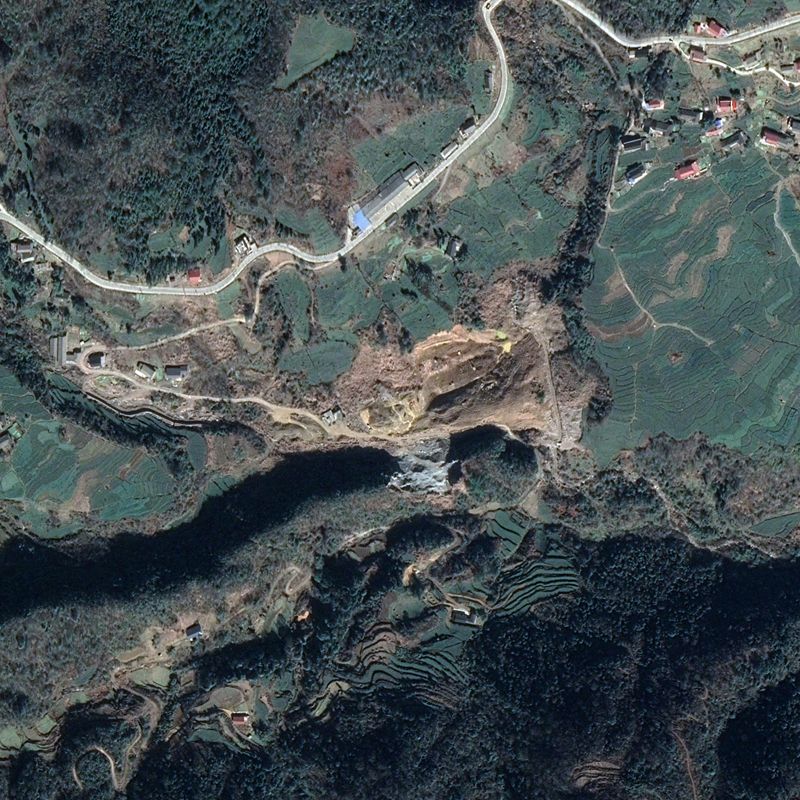}
      \vspace{1mm}
      \parbox{\textwidth}{\centering
      GVLM pre\cite{zhang2023cross}\\[-2pt]
      {\scriptsize landslide segmentation}
      }
    \end{minipage}%
  }
  \!
  \subfloat{%
    \begin{minipage}[b]{0.192\textwidth}
      \centering
      \includegraphics[width=\textwidth]{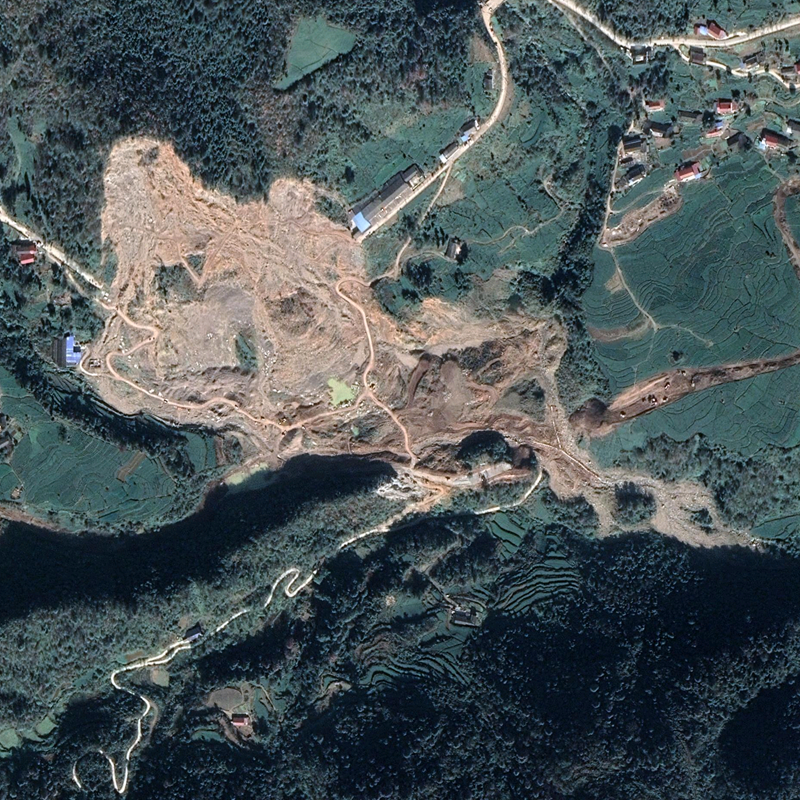}
      \vspace{1mm}
      \parbox{\textwidth}{\centering
      GVLM post\cite{zhang2023cross}\\[-2pt]
      {\scriptsize landslide segmentation}
      }
    \end{minipage}%
  }

\vspace{-4mm}
  \subfloat{%
    \label{fig:source-distr}
    \begin{minipage}[b]{0.192\textwidth}
      \centering
      \includegraphics[width=\textwidth]{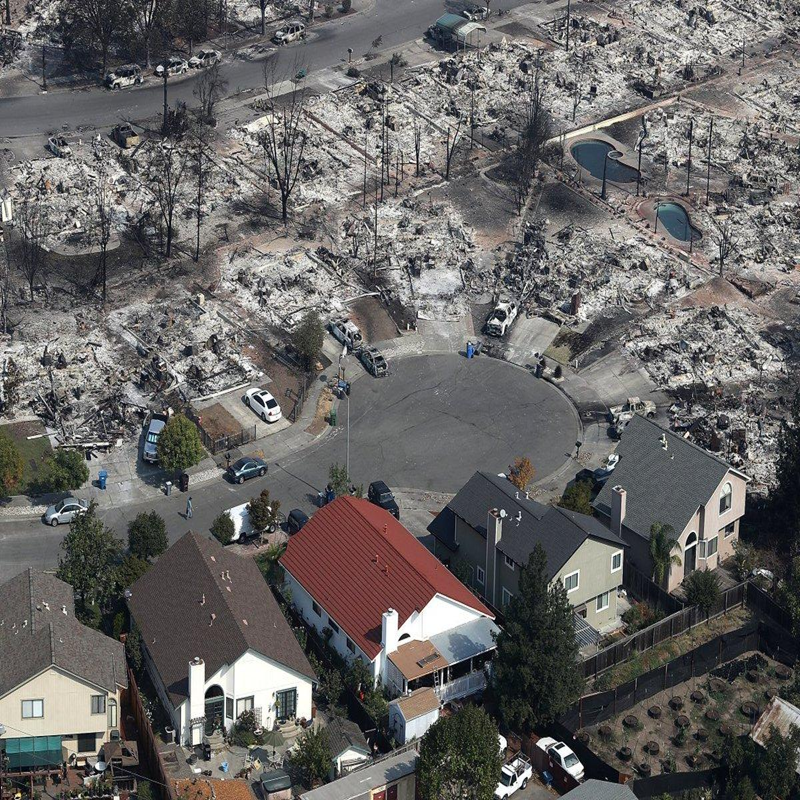}
      \vspace{1mm}
      \parbox{\textwidth}{\centering
      CrisisMMD\cite{alam2018crisismmd}\\[-2pt]
      {\scriptsize disaster classification}
      }
    \end{minipage}%
  }
  \!
  \subfloat{%
    \begin{minipage}[b]{0.192\textwidth}
      \centering
      \includegraphics[width=\textwidth]{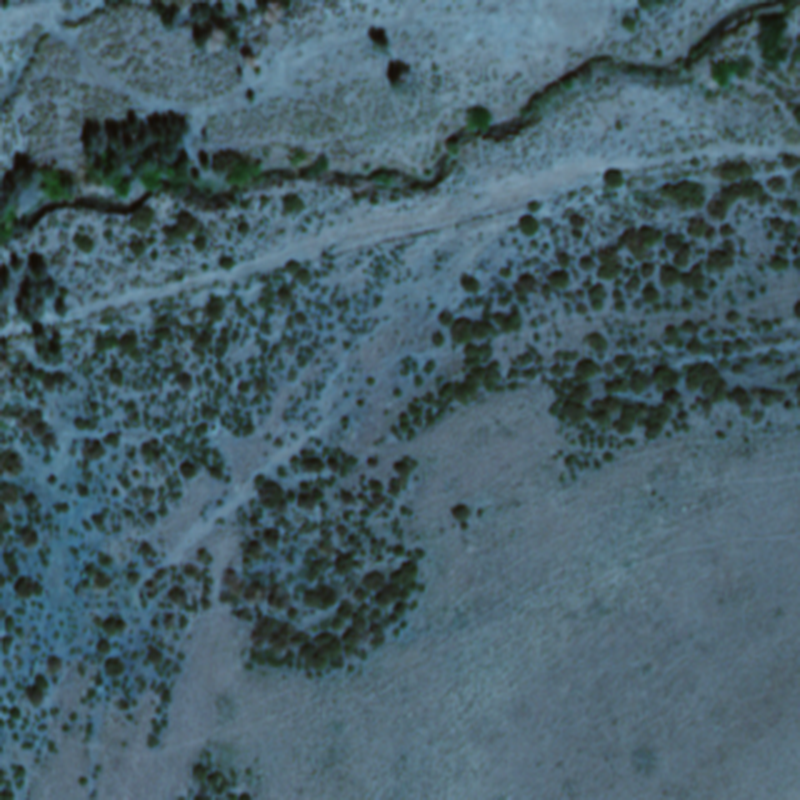}
      \vspace{1mm}
      \parbox{\textwidth}{\centering
      FireRisk\cite{shen2023firerisk}\\[-2pt]
      {\scriptsize fire risk assessment}
      }
    \end{minipage}%
  }
  \!
  \subfloat{%
    \begin{minipage}[b]{0.192\textwidth}
      \centering
      \includegraphics[width=\textwidth]{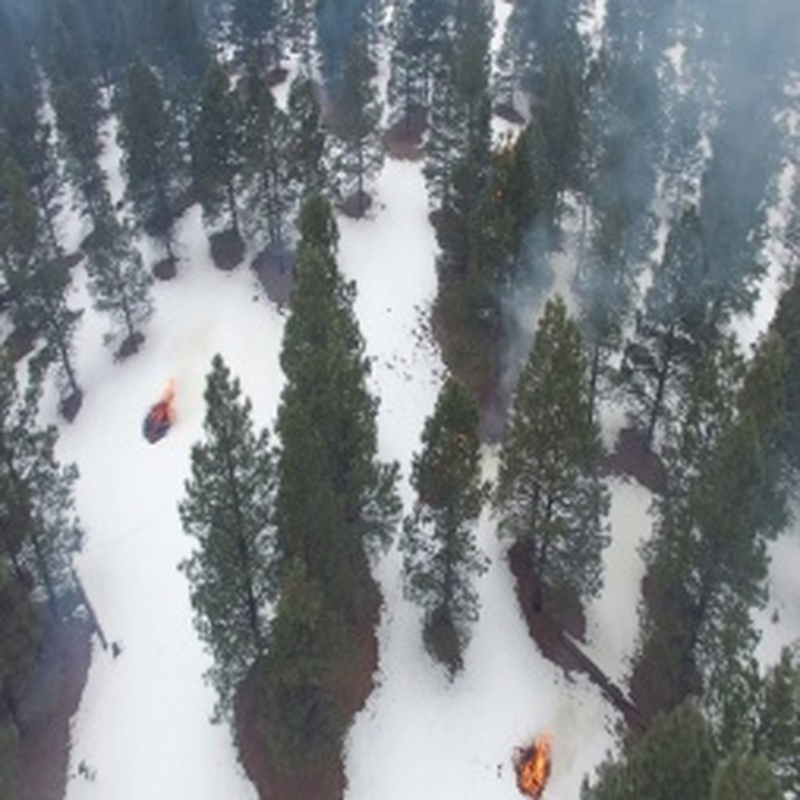}
      \vspace{1mm}
      \parbox{\textwidth}{\centering
      FLAME\cite{shamsoshoara2021aerial}\\[-2pt]
      {\scriptsize fire detection}
      }
    \end{minipage}%
  }
  \!
  \subfloat{%
    \begin{minipage}[b]{0.192\textwidth}
      \centering
      \includegraphics[width=\textwidth]{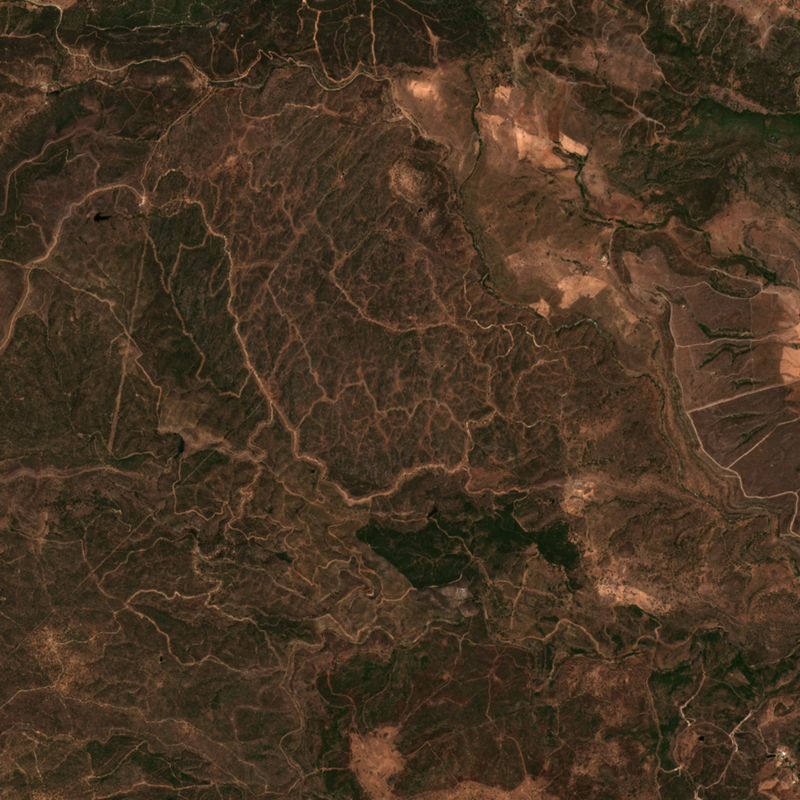}
      \vspace{1mm}
      \parbox{\textwidth}{\centering
      SBDA pre\cite{colomba2022dataset}\\[-2pt]
      {\scriptsize burned area segmentation}
      }
    \end{minipage}%
  }
  \!
  \subfloat{%
    \begin{minipage}[b]{0.192\textwidth}
      \centering
      \includegraphics[width=\textwidth]{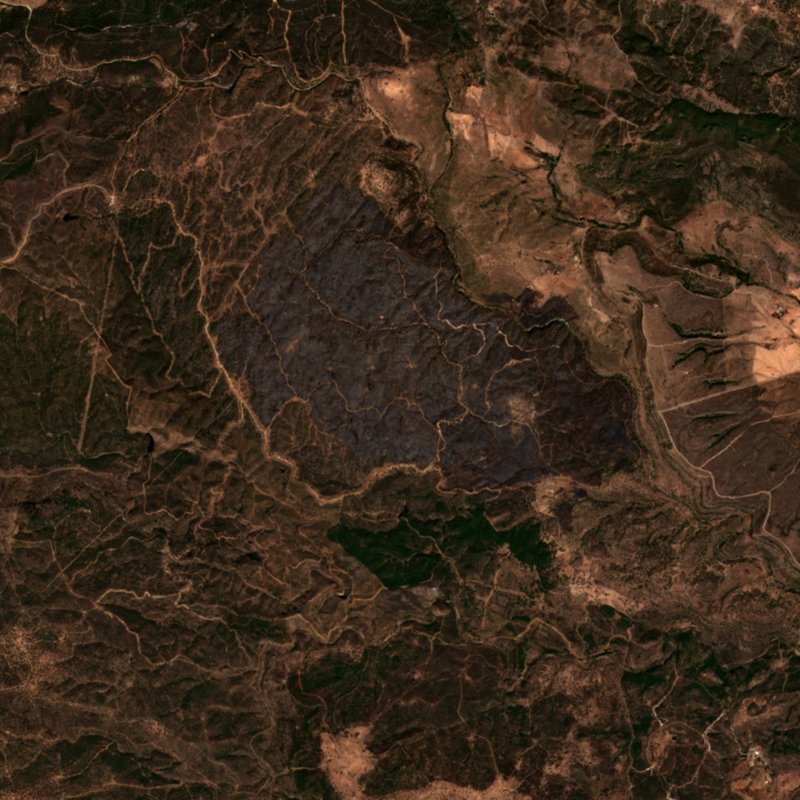}
      \vspace{1mm}
      \parbox{\textwidth}{\centering
      SBDA post\cite{colomba2022dataset}\\[-2pt]
      {\scriptsize burned area segmentation}
      }
    \end{minipage}%
  }

\vspace{-4mm}
  \subfloat{%
    \begin{minipage}[b]{0.192\textwidth}
      \centering
      \includegraphics[width=\textwidth]{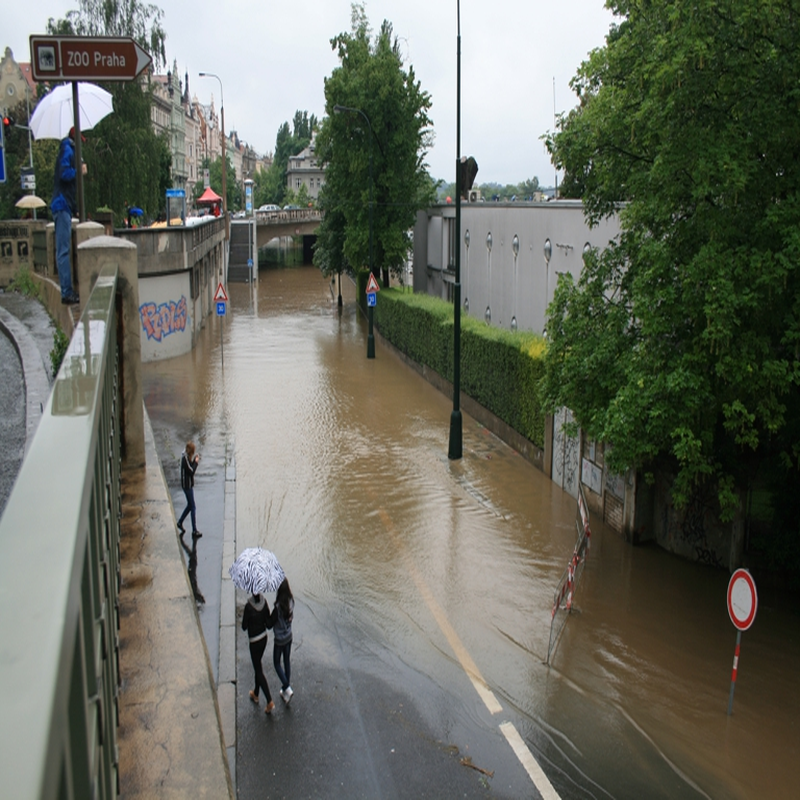}
      \vspace{1mm}
      \parbox{\textwidth}{\centering
      European Floods\cite{barz2019enhancing}\\[-2pt]
      {\scriptsize flood detection}
      }
    \end{minipage}%
  }
  \!
  \subfloat{%
    \begin{minipage}[b]{0.192\textwidth}
      \centering
      \includegraphics[width=\textwidth]{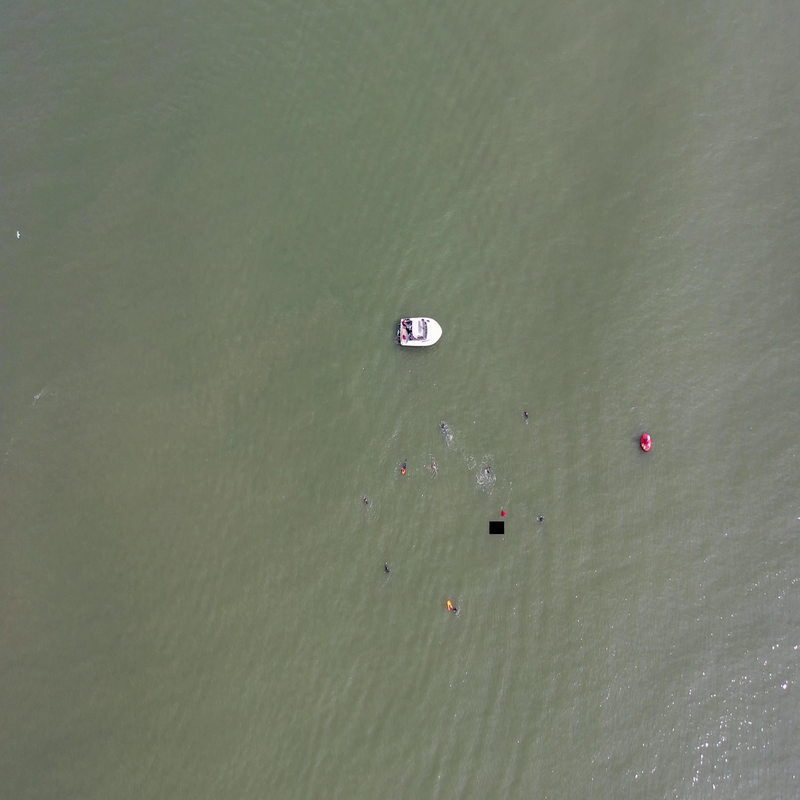}
      \vspace{1mm}
      \parbox{\textwidth}{\centering
      SeaDronesSee\cite{varga2022seadronessee}\\[-2pt]
      {\scriptsize maritime object detection}
      }
    \end{minipage}%
  }
  \!
  \subfloat{%
    \begin{minipage}[b]{0.192\textwidth}
      \centering
      \includegraphics[width=\textwidth]{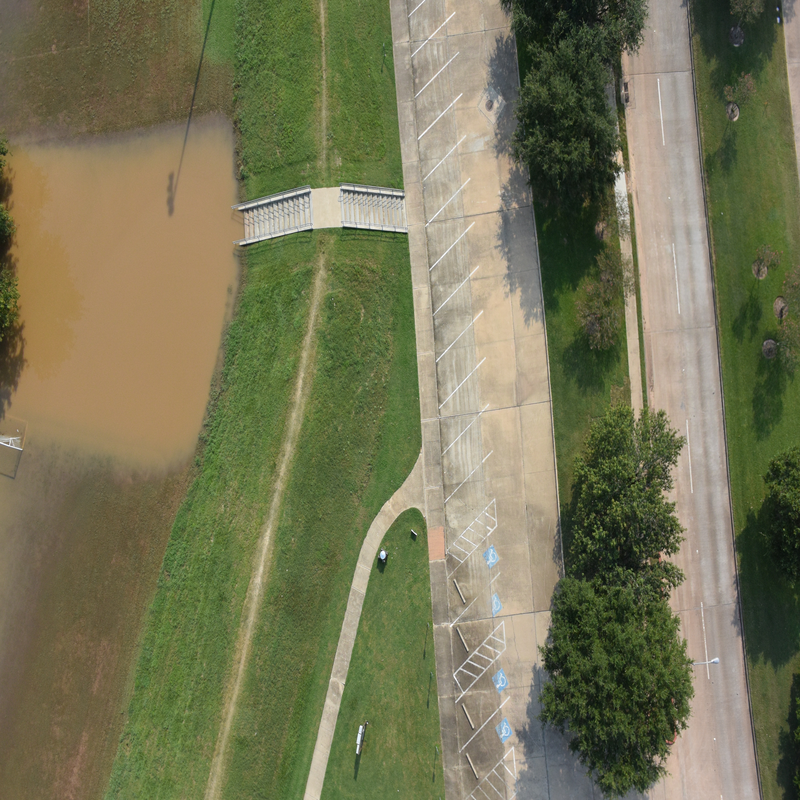}
      \vspace{1mm}
      \parbox{\textwidth}{\centering
      FloodNet\cite{rahnemoonfar2021floodnet}\\[-2pt]
      {\scriptsize flood detection}
      }
    \end{minipage}%
  }
  \!
  \subfloat{%
    \begin{minipage}[b]{0.192\textwidth}
      \centering
      \includegraphics[width=\textwidth]{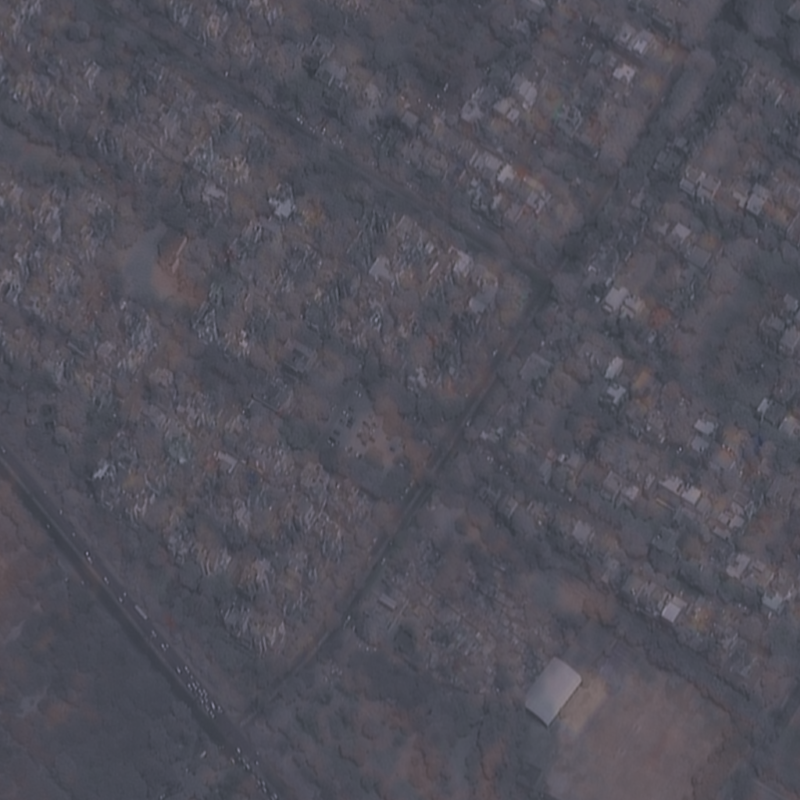}
      \vspace{1mm}
      \parbox{\textwidth}{\centering
      S2 Looking pre\cite{shen2021s2looking}\\[-2pt]
      {\scriptsize building change detection}
      }
    \end{minipage}%
  }
  \!
  \subfloat{%
    \begin{minipage}[b]{0.192\textwidth}
      \centering
      \includegraphics[width=\textwidth]{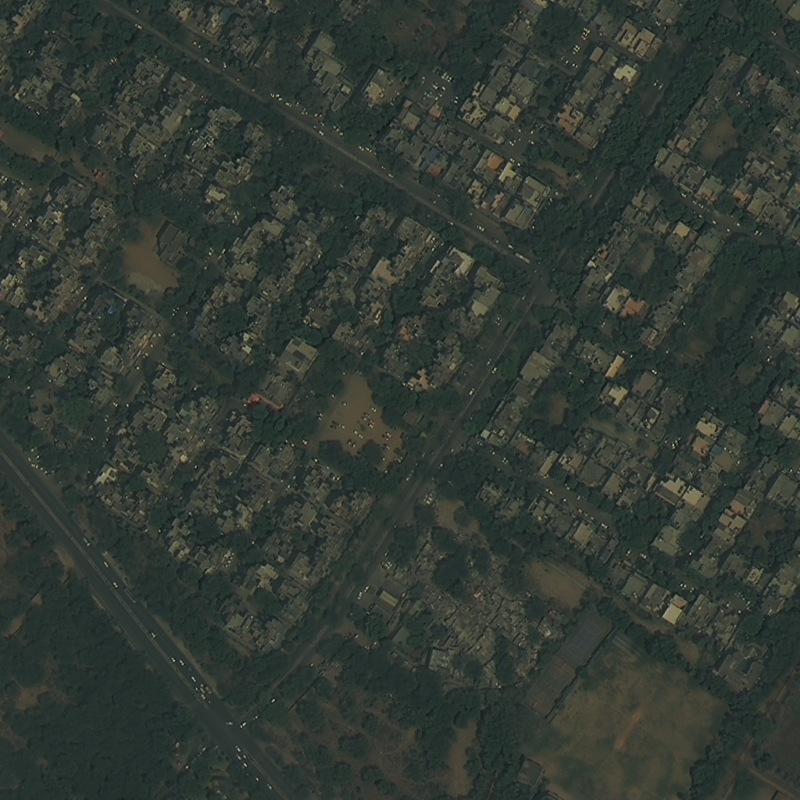}
      \vspace{1mm}
      \parbox{\textwidth}{\centering
      S2 Looking post\cite{shen2021s2looking}\\[-2pt]
      {\scriptsize building change detection}
      }
    \end{minipage}%
  }

\vspace{-1mm}
\caption{Showcasing various dataset samples. Sample images are visualized aiming at maximizing the variability in disaster types. Dataset name and corresponding task is provided below each sample image. Images are scaled to fit to the figure.}
\label{fig:showcase_datasets}
\end{figure*}
}

\noindent \textbf{Search strategy:} The literature and dataset search was conducted across multiple sources to ensure broad coverage. Academic portals and indexing platforms are used including Web of Science\footnote{https://www.webofscience.com/wos/}, Google Scholar\footnote{https://scholar.google.com/}, and ResearchGate\footnote{https://www.researchgate.net/}. General-purpose search engines (Google\footnote{https://www.google.com/}, Yahoo\footnote{https://search.yahoo.com/}) were used to identify additional datasets not formally described in academic publications. Furthermore, domain-specific repositories and research platforms including arXiv\footnote{https://arxiv.org/}, OpenML\footnote{https://www.openml.org/}, and Papers With Code\footnote{https://paperswithcode.com/} were explored. Dedicated disaster management and humanitarian data portals such as the Global Disaster Alert and Coordination System (GDACS)\footnote{https://www.gdacs.org/}, the Humanitarian Data Exchange (HDX)\footnote{https://data.humdata.org/}, and UNDRR were also systematically queried.
\noindent \textbf{Keywords and query formulation:} Search queries combined disaster-related terms including ``disaster management'', ``disaster datasets'',``disaster mitigation'', ``disaster preparedness'', ``disaster response'', ``disaster recovery'' with AI- and data-related terms (e.g., ``machine learning'', ``deep learning'', ``computer vision'', ``remote sensing'', ``satellite imagery'', ``UAV imagery''). These were further refined by specifying disaster types (e.g., ``flood'', ``earthquake'', ``wildfire'', ``tsunami'') and data modalities (e.g., ``images'', ``text'', ``sensor data'').
\noindent \textbf{Inclusion and exclusion criteria:} Datasets were included if they (i) are accessible, (ii) are relevant to disaster management or emergency response, and (iii) contain data modalities suitable for machine learning or deep learning applications, particularly vision and remote sensing. Datasets lacking sufficient documentation or relevance to disaster-related tasks were excluded. Redundant datasets and non-disaster-specific benchmarks were also omitted.
\noindent \textbf{Dataset screening and analysis:} The filtered datasets were collected for detailed analysis using a set of evaluation criteria, including data quality, accessibility, disaster type coverage, data modality, annotation or labeling scheme, dataset size, and supported machine learning tasks. The selected datasets were then categorized according to disaster type, data modality, and application domain.

In light of these criteria, we collected and listed 110 datasets in this survey.

\subsection{Comprehensive Overview of Datasets}
Table~\ref{tabl:all_datasets_together} presents a comprehensive survey of \textbf{110 disaster-related datasets} to give an overview of current disaster management data landscape. For each entry, we list: the dataset name, corresponding disaster-management stage (belonging to whether mitigation, preparedness, response or recovery stage), disaster type(s) as categories or classes, acquisition source, data modality (e.g., optical, SAR, multispectral), total number of samples in the dataset, data split percentage, included annotation types (labels, masks, bounding boxes), spatial resolution, original dataset purpose (task), and temporal coverage (pre-, during-, post-disaster or multi-temporal). The {\it stages} column categorizes datasets across the four stages of disaster management. Furthermore, the table includes metadata regarding resolution and modality of the surveyed datasets. The {\it disaster} column lists the existing disaster types in each dataset. While some datasets only focus on a single type or a few types of disasters such as flood, earthquake, tornado, tsunami, volcanic eruption, hurricane, wildfire, the category of "multiple disasters" is used for the datasets containing larger variety of disaster types, which did not fit into the table to list them individually. The acquisition source is listed under {\it source} column summarizing where and/or how the dataset is collected. Satellite (including Sentinel, Landsat, and Maxar/DigitalGlobe) is the most common source type. Other sources include curated/internet collections (e.g., Flickr, Twitter), UAV-imagery and simulation/generated data. {\it Modality} column lists the provided sensor or data types. "Optical" (RGB), "Synthetic Aperture Radar" (SAR), "infrared", "text", as well as "several bands". Most satellite imagery is available in high-resolution optical imagery and in some cases, this is accompanied by other several bands. Internet and other sources based data is most available as optical (visible) imagery and in some cases accompanied by other modalities such as text. The number of data samples, expressed in terms of images (or image pairs), video frames, text-image pairs in a dataset is presented in \textit{\# of samples} column. The {\it annotation type} indicates the format of the ground truth provided. This includes "labels" for classification, "masks" for segmentation, "bounding boxes" for detection, "text" in conjunction with any of the above for tasks such as sentiment analysis.  \textit{Spatial temporal (span)} column indicates the timeline of the data relative to the disaster. "pre" (before the event), "post" (after the event), and "during" or combination of these (pre-post) pairs in typically image format. There are some datasets providing the data in video format as well, and if video format is provided, it is listed under the "modality" column. 

Tasks column, lists what kind of disaster is contained in the dataset, under many disaster types, there might be different applications and different types of tasks focused on by each dataset. Those specific tasks are listed under the {\it Task} column. Vision-based disaster datasets can be applied in various tasks within the field of disaster management. These tasks can be broadly categorized as damage assessment, disaster detection, environmental monitoring, and other application-specific hazard identification tasks referred to as ``other tasks''.  The efficacy of models in these tasks is highly dependent on the quality and richness of the training data. To boost a model's performance and its generalization capability, \textit{damage detection and assessment}, with the highest dataset count (approx. 32 datasets in this survey), is a critical task in disaster management which involves detecting, identifying and quantifying the extent of damage from disasters. Accurate and timely damage assessment is crucial for effective response actions and prioritizing recovery efforts. The high frequency of datasets dedicated to this task highlights its importance in post-disaster scenarios. \textit{Disaster detection and classification} focuses on identifying the occurrence of disasters in images and categorizing them into relevant classes such as earthquakes, floods, and hurricanes. The datasets in this category are designed to improve the accuracy of automated detection systems for various types of disasters. Inevitably floods are among the most common natural disasters and their detection can prevent significant loss of life and property. \textit{Flood detection and mapping} enable the development of models that can predict flood occurrences, monitor water levels, and map the extent of flooded areas. \textit{Fire detection and segmentation} involves datasets for systems that can detect fires, monitor or map their extent. \textit{Building detection and segmentation} focuses on identifying and mapping buildings in disaster-affected areas. This information is useful in assessing the impact on infrastructure and planning reconstruction. Accurate building maps are also used in urban planning and development to enhance the resilience of cities against future disasters. Another significant task is \textit{Landslide detection and mapping} that addresses the identification and monitoring of landslides. Datasets in this category are used to develop models that can predict landslide occurrences and monitor affected areas. \textit{Maritime detection} included datasets built for monitoring and detecting water-related disasters such as oil spills and other open-water related accidents. Several other tasks included landcover classification, change detection, risk assessment, road extraction, waste identification, extreme weather forecasting, hurricane intensity prediction, object (people) localization, victim detection, and volcanic stage classification. Each of these tasks directly or indirectly addresses specific challenges present in its respective study and analysis of disasters, highlighting the wide area of applications in disaster management. In addition, it is important to note that each task demands a customized approach, and some tasks, such as assessing damage in disaster-affected areas, receive more attention from researchers due to their critical importance and relevance to computer vision. This emphasis on task-specific solutions underscores the complexity and nuance inherent in addressing the multitude of challenges of disaster management.

\begin{figure}[!t]
    \centering
    \includegraphics[width=\linewidth]{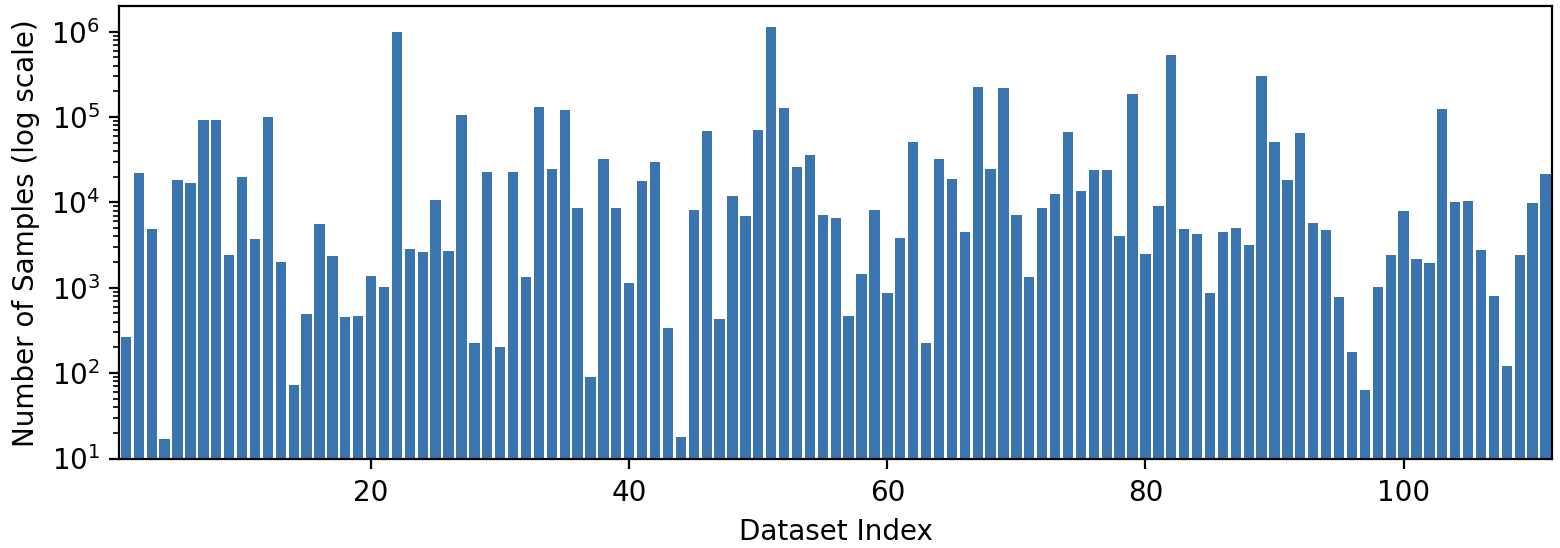}
    \caption{Total number of samples (log-scaled) vs. dataset index is visualized. The dataset index represents the dataset number in Table \ref{tabl:all_datasets_together}. }
    \label{fig:samples_distribution}
\end{figure}

\begin{figure}[!t]
    \centering
    \includegraphics[width=0.85\linewidth]{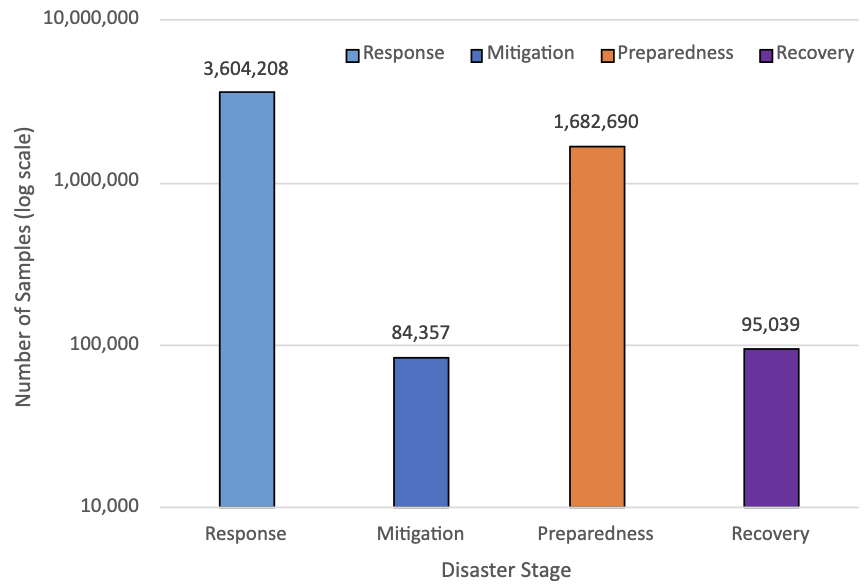}
    \caption{Total number of samples (log-scaled) per disaster stage over the entire 110 datasets are listed.}
    \label{fig:stage_samples}
\end{figure}

Figure~\ref{fig:combined-findings}(a) shows distribution of datasets based on the management phase. 67.95\% of the datasets target the \emph{response}, while a 24.34\% addresses the \emph{preparedness} stage. Datasets for  \emph{mitigation} and \emph{recovery} are each 3.85\%. Furthermore, Figure \ref{fig:stage_samples} shows the total number of images available for response, mitigation, preparation and recovery phases over the entire 110 datasets. Figure~\ref{fig:combined-findings}(b) shows main data sources from which the datasets were collected or curated from. \emph{Satellite} data is the dominant source with 44.87\%, followed by \emph{curated} collections at 23.08\%. Datasets from the \emph{internet} cover 16.67\%, while \emph{UAV}-collected datasets form 14.01\%. Remaining datasets cover simulations forming approx. ~1.28\%. In Figure~\ref{fig:combined-findings}(c), we summarized datasets into the following disaster groups: earthquake, landslide, hurricane, structural, fire, flood/tsunami, and multiple. \emph{Multiple} (datasets including multiple disasters) are 30.77\%, \emph{flood/tsunami} covers 32.05\% of the 110 datasets, followed by \emph{fire} at 15.38\%, \emph{infrastructural} at 10.26\%, and \emph{hurricane}, \emph{earthquake}, and \emph{landslide} with 2.56\%, 3.85\%, and 5.13\%, respectively. Dataset-modality distribution, shown in Figure~\ref{fig:combined-findings}(d), highlights distribution of the dataset modalities. \emph{Optical} imagery for standard visible images dominates with 69.23\%; \emph{SAR} 12.82\%, \emph{several bands} (multispectral/hyperspectral\footnote{https://pro.arcgis.com/en/pro-app/latest/help/data/imagery/satellite-sensor-raster-types.htm}; 12-bands for Sentinel-2) 7.69\%, \emph{optical and IR} 8.97\%, and \emph{optical and text} 1.28\%. In Figure~\ref{fig:combined-findings}(e), we provide the distribution of datasets based on the annotation type. \emph{masks and labels} for segmentation tasks cover 48.75\%; class \emph{labels} alone account for 35\%; \emph{bounding boxes and labels} comprise 15\%. \emph{depth maps} are the most scarce annotation types with 1.25\% in the studied datasets. Please note that these reported numeric values are individually rounded to two decimal places for simplicity. Figure \ref{fig:samples_distribution} provides total number of samples contained by each dataset in Table \ref{tabl:all_datasets_together}. Representative references for dataset examples and modality-specific collections are given throughout the text (e.g., \cite{gupta2019creating, shamsoshoara2021aerial, demir2018deepglobe, bonafilia2020sen1floods11, alam2018crisismmd, weber2020eccv, kikaki2022marida}).


{
\begin{figure*}[t]
  \centering
  \subfloat[t][{\footnotesize {\bf Per-Phase Dataset Distribution}}]{%
    \label{fig:source-distr}
      \begin{minipage}[b]{0.32\textwidth}
        \centering
        \setlength{\fboxsep}{1pt} 
        \setlength{\fboxrule}{1pt} 
        \fcolorbox{blue}{white}{
        \includegraphics[width=.982\textwidth]{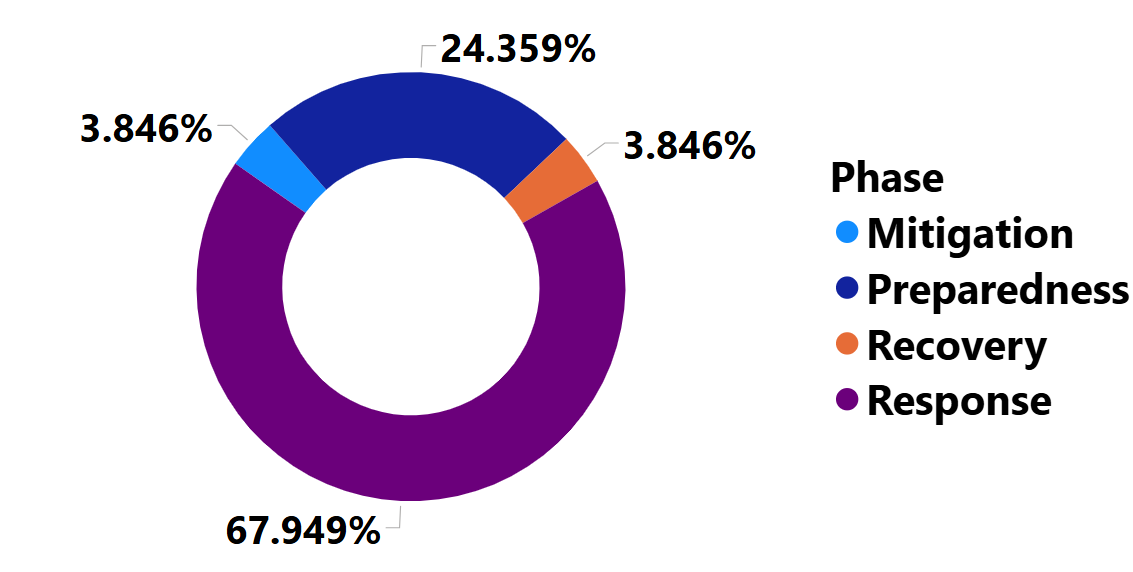}
        }
      \end{minipage}%
  }
    \hspace{1mm}
  \subfloat[t][{\footnotesize {\bf Data Source Distribution}}]{%
      \begin{minipage}[b]{0.32\textwidth}
        \centering
        \setlength{\fboxsep}{1pt} 
        \setlength{\fboxrule}{1pt}
        \fcolorbox{blue}{white}{
        \includegraphics[width=.987\textwidth]{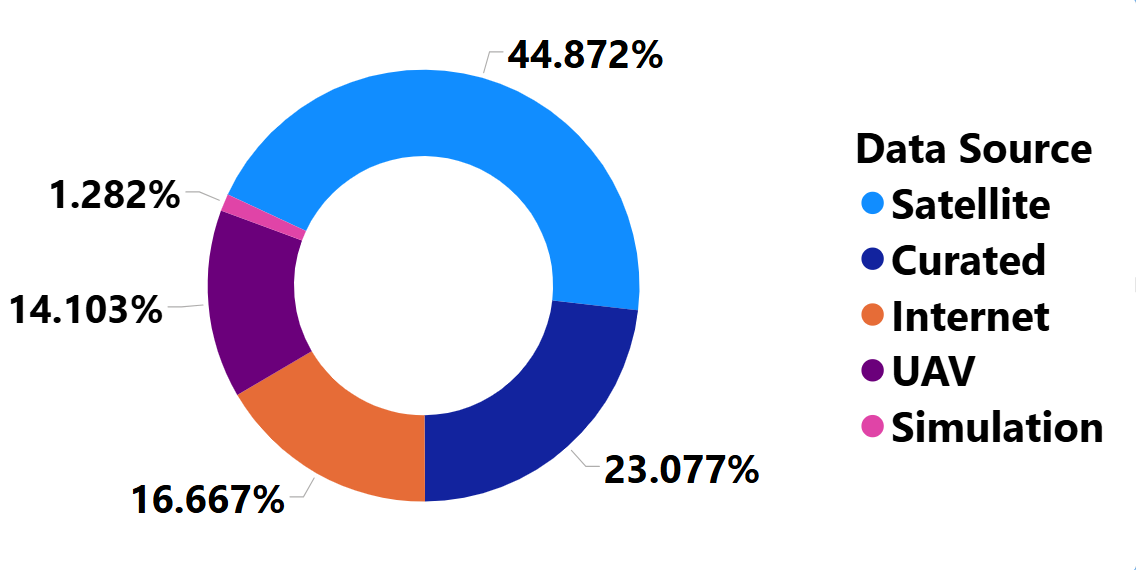}
        }
      \end{minipage}%
  }
  \subfloat[t][{\footnotesize {\bf Dataset Disaster Distribution }}]{%
      \begin{minipage}[b]{0.32\textwidth}
        \centering
        \setlength{\fboxsep}{1pt} 
        \setlength{\fboxrule}{1pt}
        \fcolorbox{blue}{white}{
        \includegraphics[width=0.986\textwidth]{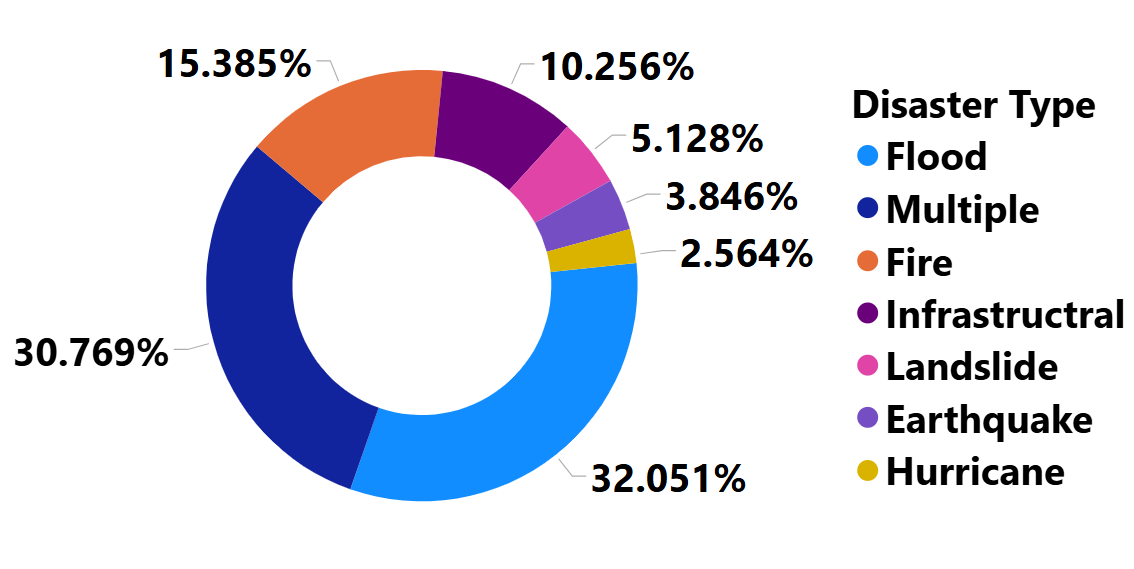}
        }
      \end{minipage}%
  }
  \vspace{-2mm}
  \subfloat[t][{\footnotesize {\bf Dataset Modality Distribution}}]{%
      \begin{minipage}[b]{0.34\textwidth}
        \centering
        \setlength{\fboxsep}{1pt} 
        \setlength{\fboxrule}{1pt}
        \fcolorbox{blue}{white}{
        \includegraphics[width=0.961\textwidth]{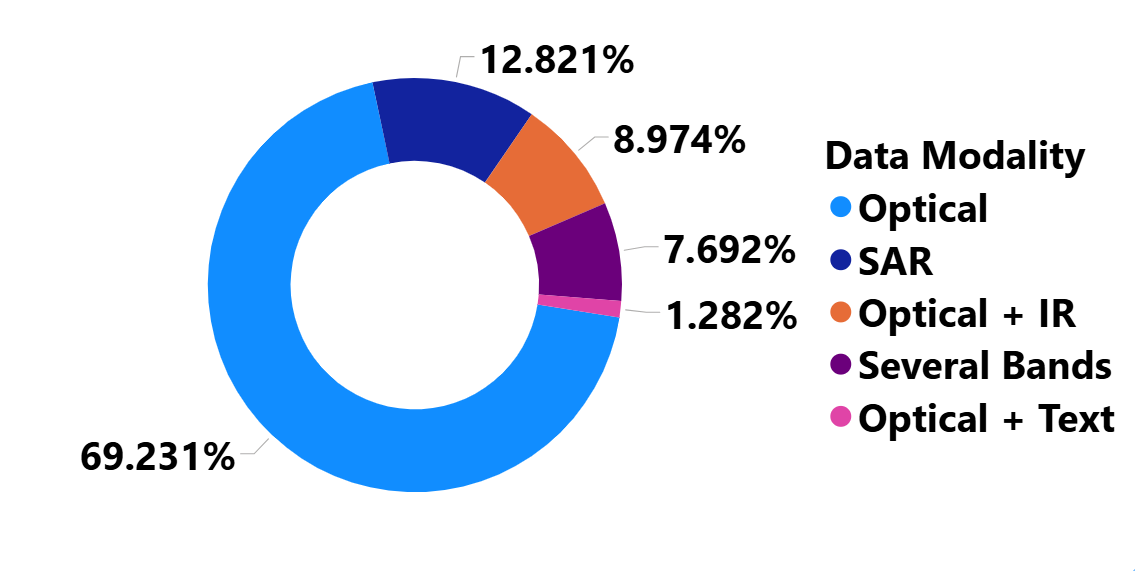}
        }
      \end{minipage}%
  }
  \subfloat[t][{\footnotesize {\bf Annotation Format Distribution}}]{%
      \begin{minipage}[b]{0.34\textwidth}
        \centering
        \setlength{\fboxsep}{1pt} 
        \setlength{\fboxrule}{1pt}
        \fcolorbox{blue}{white}{
        \includegraphics[width=0.962\textwidth]{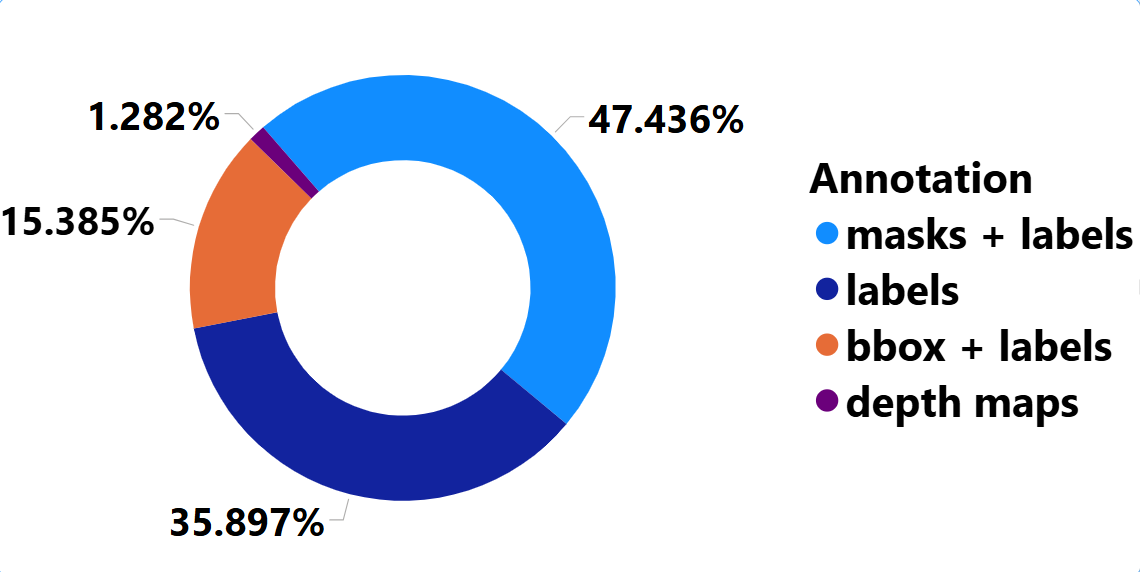}
        }
      \end{minipage}%
     
  }
  \caption{Dataset distributions are visualized over the listed 110 datasets in Table \ref{tabl:all_datasets_together} based on: (a) disaster management phases, (b) data sources, (c) disaster types, (d) data modalities, and (e) annotation formats.}
    \label{fig:combined-findings}%
\end{figure*}
}

%
%
\begin{figure}[!t]
    \centering
    \includegraphics[width=1.\linewidth]{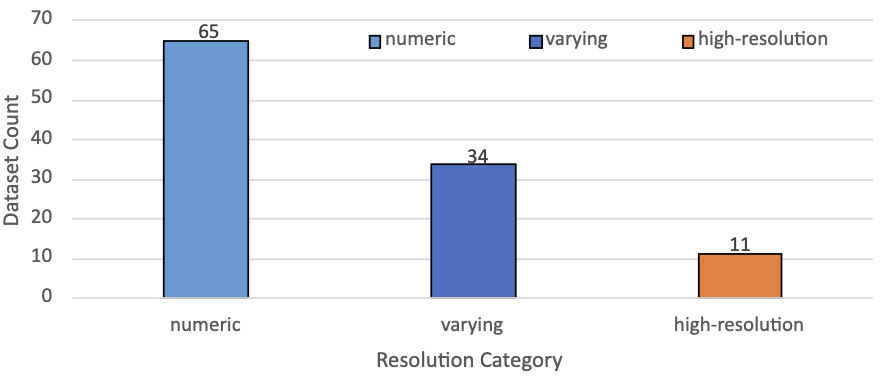}
    \caption{Distribution of resolution reporting styles in Table II. In the figure, “numeric” denotes fixed (and uniform) resolution values (e.g. 512x512 pixels), “varying” denotes a range of (non-uniform) resolutions, and “high-resolution” denotes significantly higher average pixel counts (e.g. 4000x4000 pixels).}
    \label{fig:resolution_distribution}
\end{figure}

In Table \ref{tabl:all_datasets_together}, image resolutions are reported using different levels of specificity and consistency. We group these reporting styles into three categories: numeric, varying, and high-resolution in Figure \ref{fig:resolution_distribution}. {\it Numeric} refers to datasets that provide a fixed and explicitly stated image resolution. These samples share a uniform image size, meaning every image has the same pixel dimensions (e.g., each sample is 512x512 pixels). In contrast, {\it varying} refers to datasets that contain samples with non-uniform image sizes, meaning images differ in pixel dimensions across the dataset. For example, one image may be 512x512 while another may be 1024x1024 pixels. {\it High-resolution} refers to datasets having substantially larger images on average, often in the multi-megapixel range (e.g., 4000x4000 pixels), which provides richer spatial detail and fine-grained visual analysis. Figure \ref{fig:samples_distribution}  demonstrates total number of provided samples (numbers are shown in log-scaled format) in each dataset vs. dataset index based on their index number in Table \ref{tabl:all_datasets_together}.

\subsection{Analysis of Trends}
Using the provided information in Table~\ref{tabl:all_datasets_together}, Figure~\ref{fig:combined-findings} visualizes dominant patterns across our findings.

\noindent \textbf{Phase distribution (Fig.~\ref{fig:combined-findings}a):} The distribution is strongly biased toward the \emph{response} phase (67.95\%), while \emph{preparedness} accounts for 24.36\% and \emph{mitigation} and \emph{recovery} each represent 3.85\%. This indicates a significant focus on immediate, post-impact tasks (e.g., damage mapping) rather than long-term mitigation or reconstruction in the existing datasets.

Vision-based {\it mitigation} datasets \cite{requena2020earthnet2021, bonafilia2020sen1floods11, amin2021earthquake}, for this phase, mainly consist of aerial imagery and involve activities such as monitoring \cite{mansour2020monitoring, alam2020using, karra2021global} and analyzing structural or environmental changes, such as detecting deforestation \cite{reiche2015fusing, john2022attention}, to take necessary steps toward mitigating future disasters. While imagery data is widely used for the mitigation stage, other modalities have also been explored. Gao et al. \cite{gao2024exploring} investigated the management of flood risk in coastal areas by measuring the first-floor elevation (FFE) of buildings, a crucial factor in flood risk, using deep learning and Google Street View data. {\it Preparedness} involves planning and training to respond to a likely (imminent) disaster before it occurs. Information about infrastructure maps (roads), building footprints, and population density maps \cite{bountos2022hephaestus, van2018spacenet, demir2018deepglobe, huot2022next} and their respective models is crucial for planning evacuation routes, setting-up emergency shelters and allocating resources in times of disaster. {\it Response} involves taking immediate actions to save lives and minimize damage during a disaster. Real-time satellite or drone imagery \cite{gupta2019creating, shamsoshoara2021aerial, rahnemoonfar2021floodnet, fujita2017damage, akshya2019hybrid} of areas affected by disasters provides situational awareness for {\it Response}. Other modalities, i.e., social media data \cite{alam2018crisismmd, barz2021finding, alam2020deep, chaudhary2020water, alam2023medic, benjamin2017multimedia} have been studied. Weber et al. \cite{weber2020eccv} introduced the {\it Incidents Dataset}, containing over \textit{446,000} positive and over \textit{697,464} negative annotated images collected from Flickr and Twitter covering \textit{43} incident types. Using clues extracted from these data helps prioritize rescue efforts. {\it Recovery} covers restoring infrastructure and providing aid to affected communities. Datasets \cite{colomba2022dataset, kikaki2022marida, zhang2023assessment} for this phase include tracking the progression of recovery efforts, and damage assessment reports can be used to prioritize reconstruction projects and allocate funding. The distribution given in \emph{Figure \ref{fig:combined-findings}a} reveals that vision solutions are more applied in the response stage.

\noindent \textbf{Data sources (Fig.~\ref{fig:combined-findings}b):} Satellite-derived datasets are the dominant source (44.87\%), followed by curated academic collections (23.08\%), internet/social-media (16.67\%), UAV imagery (14.10\%) and simulations (~1.28\%). The satellite imagery comes in high-resolution spatially and comes with metadata which makes it suitable for tasks requiring high resolution images and/or fusing different modalities. However, satellite imagery is costly and not easy to find in high temporal resolution. On the other hand, the internet and social media data is abundant and easy to scrape but comes with noisy labels and often with poor or in non-standardized quality.

The collection of disaster datasets is challenging, as data may come from various sources such as satellite imagery \cite{chen2018benchmark, hansch2022spacenet, huot2022next}, social-media \cite{barz2021finding, alam2018crisismmd}, and ground sensors, leading to variations in data formats, modality, resolutions, and quality. The design of an efficient machine (or deep) learning model, typically, depends on the kind of data it will receive for training. The challenges associated with data collection initially depend on the source platform. \emph{Figure \ref{fig:combined-findings}b} illustrates the distribution of the data sources over the 110 datasets summarized in Table \ref{tabl:all_datasets_together}. Satellite data may come from proprietary providers. Social media platforms such as ``Twitter'' (which has become ``X'') serve as data sources (image and text based) for especially rapid detection applications, provided by the community through exposed application program interfaces (API) \cite{layek2019detection, chaudhary2020water, alam2023medic}. Furthermore, the internet is massively \cite{kyrkou2019deep, jadon2019firenet, sesver2022vidi} utilized to collect data of different formats and modalities through news portals and crowd-sourcing platforms, or simply by crawling the internet. Data collected from such sources comes in various formats (i.e., image, text, etc.) and is noisy and as a result requires intensive preprocessing. Techniques such as hard negative mining \cite{dong2017class} are also proposed to increase the variability of the dataset.

\noindent \textbf{Formats and resolutions: }In our observations, the formats greatly varied from dataset to dataset and a main driving reason was the data source. Satellite and airborne imagery often provide high-definition optical, multispectral bands \cite{dong2023enteromorpha, weir2019spacenet, rudner2019multi3net}, or SAR imagery \cite{bonafilia2020sen1floods11, dotel2020disaster}. The dataset may have varying resolutions based on the used sensors and their parameters \cite{shamsoshoara2021aerial, jadon2019firenet, bai2021end}, thus providing richer or poorer feature representations. Satellite imagery can provide high- or very high-definition ranging from a submeter (30-50 cm) to 30-50 meters. That is, a $2048 \times 2048$ typical satellite image size would cover a space of tens of kilometers. Satellite imagery is appropriate for more use cases, as various constellations offer different spectral bands and resolutions and high-quality data. Unlike satellite imagery, datasets built on social networks and internet images have a smaller resolution (as small as $50 \times 50$ \cite{devaraj2021novel} to $1024 \times 1024$ \cite{chen2018benchmark}), since they are mostly uploaded or taken by individuals. It is also the case that datasets from the internet and social media require longer pre-processing pipelines. In addition to those sources, data can also be generated through simulations \cite{schmidt2022climategan} for a more tailored purpose. Although simulations offer controlled and repeatable setups, in many cases, they require a domain expert with the domain knowledge. Figure \ref{fig:resolution_distribution} highlights the distribution of "numeric", "varying" and "high-resolution" datasets over the studied 110 datasets. As observed in the figure, datasets with high spatial resolution are comparatively rare.

\noindent \textbf{Disaster types (Fig.~\ref{fig:combined-findings}c):} Datasets covering \emph{flood/tsunami} disaster types lead at 32.05\%, closely followed by \emph{multiple} at 30.77\%. Fire (15.39\%) and infrastructural/building-focused datasets (10.26\%) follow; landslide datasets constitute 5\%, while hurricane and earthquake datasets are relatively scarce at 3.75\% each. The variety in disaster types, whether natural or human-caused, poses unique challenges to the availability and applicability of vision-based datasets for disaster management. Natural calamities such as earthquakes and hurricanes are typically disrupt infrastructure and hinder on-site data collection efforts at larger scales. Beyond the difficulties associated with data collection and preprocessing, such as addressing class imbalances in dataset distributions when training models for management tasks, each disaster type demands unique considerations in addition to standard vision data preprocessing protocols and routines. In cases of floods where property and infrastructure damage typically disrupt data flow and, as such, hinders accessing real-time information. Therefore, satellite imagery may serve as an alternative source of rich data. A considerable number of datasets (mainly for classification) have multiple types of disasters (floods, hurricane, earthquake, etc.) \cite{sesver2022vidi, kordopatis2019fivr, niloy2021novel}, while many others, as in \cite{shen2023firerisk, barz2019enhancing, bai2021end, munawar2021uavs}, are only specifically dedicated to one kind of disaster.

\noindent \textbf{Modalities and annotations (Fig.~\ref{fig:combined-findings}d--e):} Optical imagery dominates (69.23\%), SAR comprises 12.82\%, multispectral stacks (several bands) 7.69\%, and combined optical+IR 8.97\% (optical+text 1.28\%). Annotation-wise, pixel-level masks and labels are most common (47.44\%), class labels alone cover 35.90\%, bounding-box datasets constitute 15.39\%, and depth annotations are rare (1.28\%).

Labeling or annotation processes are an important part of useful disaster datasets for image based training since most successful, reliable and current vision models are known to be supervised algorithms requiring training labels. Labeling such disaster-specific datasets is labor intensive and time consuming, requiring domain expertise and specialized knowledge in many domains under different modalities. Annotators, even if they are experts, may interpret disaster-related features differently, leading to inconsistencies and variability in the annotations. Additionally, if only a small amount of labeled or annotated data is available, the limited availability of annotated disaster datasets can hinder the training of robust computer vision models.

\noindent These trends reveal clear couplings: satellite-sourced datasets favor mapping and segmentation tasks at larger scales, while internet/UAV collections favor classification and localized and simultaneous detection but often suffer from lower resolution and noisier labels. The predominance of optical data also indicates under-utilization of complementary modalities (e.g., SAR, thermal) in many tasks.

\subsection{Research Gaps and Future Directions}

A complete disaster management contains multiple phases as listed previously. A full and effective system focusing on all aspects of such complete disaster management system would benefit local communities, economy and infrastructure. While a variety of datasets are available for disaster-related tasks, they come with limitations that impact their usability and effectiveness. Understanding these challenges is critical for improving the selection, development, and application of datasets for specific computer vision and remote sensing tasks. Based on the comparative analysis provided above, we summarize a set of principal limitations and recommend directions for future relevant work.

\noindent \textbf{Phase bias:} A pronounced bias toward the response phase (67.95\%) limits resources for mitigation and recovery research, constraining development of prevention and long-term resilience tools. There is still need for investing in new datasets for mitigation and recovery phases and for less-represented hazards (e.g., landslides, slow-onset events) to balance the research focus.

\noindent \textbf{Modalities and Fusion Challenges:} Optical-only datasets predominate (69.23\%). Multimodal datasets that combine optical, SAR, thermal, and ground/sensor or social-media streams are scarce; this restricts robustness to adverse conditions (e.g., cloud cover, night-time) and cross-sensor generalization. While additional modalities such as SAR (useful for cloud penetration) and IR (effective for fire detection) exist, they are rarely integrated into a single dataset. The lack of multi-modal datasets creates a barrier to developing more robust and efficient fusion models. Addressing spatial and temporal registration issues in a more generic and mainstream way remains a key technical challenge for the community to enable effective multi-sensor disaster monitoring. Accessibility of such multimodal data to a broader research community also remains a significant challenge.

\noindent \textbf{Annotation consistency:} While pixel-level masks exist in many datasets, richly annotated multi-task datasets (e.g., combined segmentation, instance detection, and temporal change labels) are limited. Annotation heterogeneity and inconsistent labeling standards hamper dataset integration and model transfer.

\noindent \textbf{The Quality and Diversity Trade-off:} Existing datasets often face a trade-off between high resolution and category diversity. High-resolution satellite and UAV datasets frequently cover limited disaster scenarios such as landslides, earthquakes and floods, whereas large-scale internet-crawled datasets suffer from noisy labels and varying quality, yet cover more disasters. Future research should focus on curating diverse, high-fidelity and possibly multi-modal datasets that maintain semantic richness across multiple disaster types. 

High-resolution satellite and UAV imagery require substantial storage with sizes reaching up to Terabytes. That size alone requires significant compute resources, limiting reproducible experimentation for groups without large infrastructure. Creating datasets that explicitly combine optical, SAR, thermal, and ancillary (GIS, population, social media) layers across pre/during/post phases enables robust, real-world models. 
The variety of data modalities further complicates the usability of the dataset. Disaster datasets may include optical imagery \cite{alam2018crisismmd, pi2021detection, niloy2021novel, alam2020deep, kyrkou2019deep}, infrared \cite{shamsoshoara2021aerial, qi2020automatic, van2018spacenet}, SAR data \cite{bonafilia2020sen1floods11, dotel2020disaster, gahlot2022curating}, and multi-spectral satellite bands \cite{van2018spacenet, weir2019spacenet, kikaki2022marida}. While each modality has distinct advantages (for example, infrared for detecting fire), many datasets provide only a single modality and blending in complementary data adds an extra complexity including possible spatial or temporal registration issues \cite{ozer2024visirnet}. Dataset size and processing requirements also present their own computational challenges \cite{andreadis2020flood, chen2018benchmark, christie2018functional, montello2022mmflood}. High-resolution aerial images, while semantically rich for detailed analysis, require substantial computational resources for storage, processing, and model training. Many disaster response teams and research groups may not have access to the necessary infrastructure, restricting the practical application of these datasets.

Finally, while there are existing repositories, they are limited to certain types of datasets. However, there is no widely recognized and centralized repository that systematically curates, organizes, and maintains disaster-related vision datasets. Datasets are scattered across different research groups and organizations, making it difficult to browse or integrate multiple sources. The lack of standardized dataset formats and annotation guidelines further adds to the complexity which makes it hard to curate universal models that generalize across different disaster types.  

\section{Conclusion}

The design of resilient and adaptive disaster management systems that effectively mitigate risks and protect communities and existing infrastructure amid increasingly frequent and complex disasters has emerged as a critical challenge.

This survey aims to provide necessary tools to researchers and practitioners for rapid development of remote sensing based disaster management applications. While we acknowledge that there may always be additional and new resources, applications and datasets, we provided a comprehensive overview of computer vision and remote sensing based approaches to disaster management and response, focusing on the datasets available for research, development, and application in this critical field. Furthermore, we summarize existing limitations for possible future research directions. Researchers can use this survey as a starting point to gain rapid exposure to the relevant terms and datasets. Through exploration of various datasets, ranging from satellite and aerial imagery to ground-based sensor data, it is evident that spatial, visual or spatio-temporal techniques plays a crucial role in enhancing our capabilities for disaster preparedness, monitoring, and rapid response. The availability of diverse datasets over time (see. \emph{Figure \ref{fig:combined-findings}f}), helps develop innovative solutions for early detection, damage assessment, and resource allocation during disasters. Careful integration of recent deep learning techniques with new or available datasets improves the accuracy and efficiency of disaster management systems significantly. 

Future studies are expected to contribute significantly to existing relevant datasets and disaster management systems. Exploring novel and additional data sources, such as filtering social media feeds, crowd-sourced imagery, and UAVs, presents opportunities to enhance situational awareness and decision-making capabilities in various disaster scenarios. Collaboration between academia, industry, and government agencies will remain crucial to facilitate access to large-scale datasets and to build effective disaster management systems.

{\small
\bibliographystyle{ieee}
\bibliography{egbib}
}

\par
\section*{Authors' Bio:}

\par
  \textbf{Alain Patrick Ndigande} received his B.Sc. degree from Kocaeli University, Turkiye, in 2022; and his M.Sc. degree from Ozyegin University in 2025. He is currently a PhD student at Ozyegin University. His research interests are Artificial Intelligence, Machine Learning, Computer Vision, Cloud Computing, ML and Data systems.

\par
  \textbf{Josiah Wiggins} received his B.Sc. degree from California State Polytechnic University, Pomona, in 2025; and he is currently a M.Sc. student at California State Polytechnic University, Pomona (Cal. Poly. Pomona). His research interests include object detection, object segmentation and remote sensing applications.  
    \par
 \textbf{Sedat Ozer} received his M.Sc. degree from Univ. of Massachusetts, Dartmouth and his Ph.D. degree from Rutgers University, NJ. He has worked as a research associate in various institutions including Univ. of Virginia and Massachusetts Institute of Technology. His research interests include pattern analysis, remote sensing, object detection \& segmentation, object tracking, visual data analysis, geometric and explainable AI algorithms and explainable fusion algorithms. 
 As a recipient of TUBITAK's international outstanding research fellow and as an Assistant Professor, he worked at the Dept. of Computer Science, Ozyegin University. He is currently at the department of Electrical and Computer Engineering at California State Polytechnic University, Pomona, CA, USA.

\end{document}